%% file: DMT-Net.tex
\documentclass[sigconf]{acmart}
\usepackage{graphicx}
\usepackage{grffile}
\usepackage{balance}
\usepackage{tabularx}
\AtBeginDocument{%
  \providecommand\BibTeX{{%
    \normalfont B\kern-0.5em{\scshape i\kern-0.25em b}\kern-0.8em\TeX}}}

\copyrightyear{2021} 
\acmYear{2021} 
\setcopyright{acmcopyright}
\acmConference[MM '21]{Proceedings of the 29th ACM International Conference on Multimedia}{October 20--24, 2021}{Virtual Event, China}
\acmBooktitle{Proceedings of the 29th ACM International Conference on Multimedia (MM '21), October 20--24, 2021, Virtual Event, China}
\acmPrice{15.00}
\acmDOI{10.1145/3474085.3475331}
\acmISBN{978-1-4503-8651-7/21/10}

\begin{document}

\fancyhead{}

\title{From Synthetic to Real: Image Dehazing  Collaborating \\ with Unlabeled Real Data}



\author{Ye Liu}
\authornote{Joint first authors. $\dagger$ Corresponding author.}
\affiliation{%
  \institution{College of Intelligence and Computing, Tianjin University}
  \city{Tianjin}
  \country{China}
}
\email{liuye321@tju.edu.cn}

\author{Lei Zhu}
\authornotemark[1]
\affiliation{%
  \institution{University of Cambridge}
  \city{Cambridge}
  \country{UK}
}
\email{lz437@cam.ac.uk}

\author{Shunda Pei}
\affiliation{%
  \institution{College of Intelligence and Computing, Tianjin University}
  \city{Tianjin}
  \country{China}}
\email{peishunda@tju.edu.cn}

\author{Huazhu Fu}
\affiliation{%
  \institution{Inception Institute of Artificial Intelligence}
  \city{Abu Dhabi}
  \country{UAE}
}
\email{hzfu@ieee.org}

\author{Jing Qin}
\affiliation{%
 \institution{The Hong Kong Polytechnic University}
 \city{Hong Kong}
 \country{China}}
\email{harry.qin@polyu.edu.hk}

\author{Qing Zhang}
\affiliation{%
  \institution{Sun Yat-sen University}
  \city{Guangzhou}
  \country{China}}
\email{zhangqing.whu.cs@gmail.com}

\author{Liang Wan}
\authornotemark[2]
\affiliation{%
  \institution{Tianjin University}
  \city{Tianjin}
  \country{China}
}
\email{lwan@tju.edu.cn}

\author{Wei Feng}
\affiliation{%
  \institution{Tianjin University}
  \city{Tianjin}
  \country{China}}
\email{wfeng@ieee.org}




\begin{abstract}
Single image dehazing is a challenging task, for which the domain shift between synthetic training data and real-world testing images usually leads to degradation of existing methods. 
%
To address this issue, we propose a novel image dehazing framework collaborating with unlabeled real data.
First, we develop a disentangled image dehazing network (DID-Net), which disentangles the feature representations into three component maps, i.e. the latent haze-free image, the transmission map, and the global atmospheric light estimate, respecting the physical model of a haze process.
Our DID-Net predicts the three component maps by progressively integrating features across scales, and refines each map by passing an independent refinement network.
Then a disentangled-consistency mean-teacher network (DMT-Net) is employed to collaborate unlabeled real data for boosting single image dehazing. Specifically, we encourage the coarse predictions and refinements of each disentangled component to be consistent between the student and teacher networks by using a consistency loss on unlabeled real data.
We make comparison with 13 state-of-the-art dehazing methods on a new collected dataset (Haze4K) and two  widely-used dehazing datasets (i.e., SOTS and HazeRD), as well as on real-world hazy images.
Experimental results demonstrate that our method has obvious quantitative and qualitative improvements over the existing methods. 
\end{abstract}





\maketitle

\input{intro}

\input{rel}

\input{method}

\input{ex}


\section{Conclusion}

This work presents a disentangled-consistency mean teacher network (DMT-Net) for boosting single-image dehazing by leveraging feature disentangled learning and unlabeled real-world images.
Our key idea is to first disentangle features from input hazy photos for simultaneously predicting clean images, transmission maps, and atmospheric images, for which we develop a disentangled image dehazing network (DID-Net) following a coarse-to-fine strategy. 
Then we assign DID-Net as the student and teacher networks to impose disentangled consistency loss for leveraging additional unlabeled data. 
Experimental results on synthesized datasets and real-world photos demonstrate the effectiveness of our network, which clearly outperforms the state-of-the-art image dehazing methods.


\noindent
\textbf{Acknowledgments:} The work is supported by the National Natural Science Foundation of China (Grant No. 61902275), the research fund for The Tianjin Key Lab for Advanced Signal Processing, Civil Aviation University of China (Grant No. 2019AP-TJ01), and a grant under Innovation and Technology Fund - Midstream Research Programme for Universities (ITF-MRP) (Project no. MRP/022/20X).


%





\bibliographystyle{ACM-Reference-Format}
\balance
\bibliography{sample-base}
%

%
%

%





\end{document}

%% file: intro.tex

\section{Introduction}
\label{sec:introduction}

Hazy images usually suffer from content distorting and accuracy degrading for subsequent visual analysis.
To improve the overall scene visibility, many image dehazing methods~\cite{cheng2018semantic,galdran2018duality,Li_2018_CVPR,ren2018gated,zhang2018densely,yang2018proximal}
have been proposed to recover the latent haze-free image from the single hazy input.
The image degradation caused by the haze could be formulated by a physical model~\cite{nayar1999vision,zhang2018densely,yang2018proximal}:
\begin{equation}  \label{Eq:physical}
I =J \cdot T +A \cdot (1-T ),
\end{equation}
where $I$ is the observed hazy image, 
$J$ is the underlying haze-free image to be recovered,
$T$ is the transmission map, which represents the distance-dependent factor affecting the fraction of light that reaches the camera sensor, and $A$ is the global atmospheric light, indicating the ambient light intensity. 
Early dehazing methods~\cite{berman2017air,fattal2008single,sulami2014automatic,ancuti2013single} employed hand-crafted priors~\cite{he2011single,fattal2014dehazing,zhu2015fast} based on the statistics of clean images to estimate the transmission map $T$, and then use the physical model to recover the haze-free results. 
Recently, a lot of methods based on convolutional neural networks (CNNs) are proposed to learn the transmission map from labeled datasets~\cite{cai2016dehazenet,ren2016single,li2017all,Li2018_Access},
or directly build the mapping from input hazy images to haze-free counterparts~\cite{Li_2018_CVPR,ren2018gated,zhang2018densely,yang2018proximal,liu2019griddehazenet,qu2019enhanced,deng2019deep,Li2020_tmm}.

Although achieving superior image dehazing performances over methods based on hand-crafted priors, existing CNN-based methods suffer from several limitations.
\textbf{First}, these methods usually utilize synthesized hazy images to train networks in a supervised learning manner, and thus suffer from degraded performance in real-world hazy photos due to the domain shift between synthetic training images and real-world testing photos.
\textbf{Second}, according to the physical model of Eq.~\eqref{Eq:physical}, an input hazy image is a combination of a transmission map, a global atmospheric map, and an underlying haze-free image, showing that CNN features learned from input hazy image include several factors of the physical model.
Unfortunately, many existing methods employed such CNN features to predict only one factor (e.g., transmission map or haze-free result), hindering image dehazing performance.



To address these problems, this work develops a disentangled image dehazing framework to leverage disentangled feature learning and unlabeled real data for boosting image dehazing performance.
Specifically, we first propose a disentangled image dehazing network (DID-Net) to disentangle features at each scale into three feature
components, which are transmission-distilled features for a transmission map estimation, latent-distilled features for a latent haze-free image estimation, and light-distilled features for a global atmospheric light estimation.
After that, we progressively integrate transmission features, latent image features, and light features at adjacent scales to predict a transmission map, a haze-free image, and a global atmospheric map.
On the other hand, for integrating synthesized and real-world hazy images, we first assign DID-Net into a mean-teacher framework, and then compute a disentangled supervised loss on labeled synthesized data and a consistency loss on unlabeled real-world data to constrain the coarse predictions and refinements of the network.
By doing so, our approach achieves a superior dehazing performance over state-of-the-art methods.
The contributions of this work are:
\begin{itemize} 
	\item We present an image dehazing framework to leverage disentangled feature representations and unlabeled real-world hazy images for boosting single image dehazing.
	\item We devise a disentangled image dehazing network (DID-Net) to predict a transmission map, a latent haze-free image, and an atmospheric light map via a coarse-to-fine strategy.
	\item A disentangled-consistency mean-teacher network  (DMT-Net) is employed to collaborate the labeled synthetic data and unlabeled real data with disentangled consistency losses. 

\end{itemize}

We compare our network against 13 state-of-the-art dehazing methods on a new collected dataset, a widely-used dehazing benchmark datasets and various real-world hazy images. The experimental results demonstrate that our network outperforms state-of-the-art dehazing methods. 
\textit{Our code, trained models, and results at \url{https://github.com/liuye123321/DMT-Net}.
}

%% file: rel.tex
\section{Related Work}

\begin{figure*}
	\begin{center}
		\includegraphics[width=1.0\linewidth]{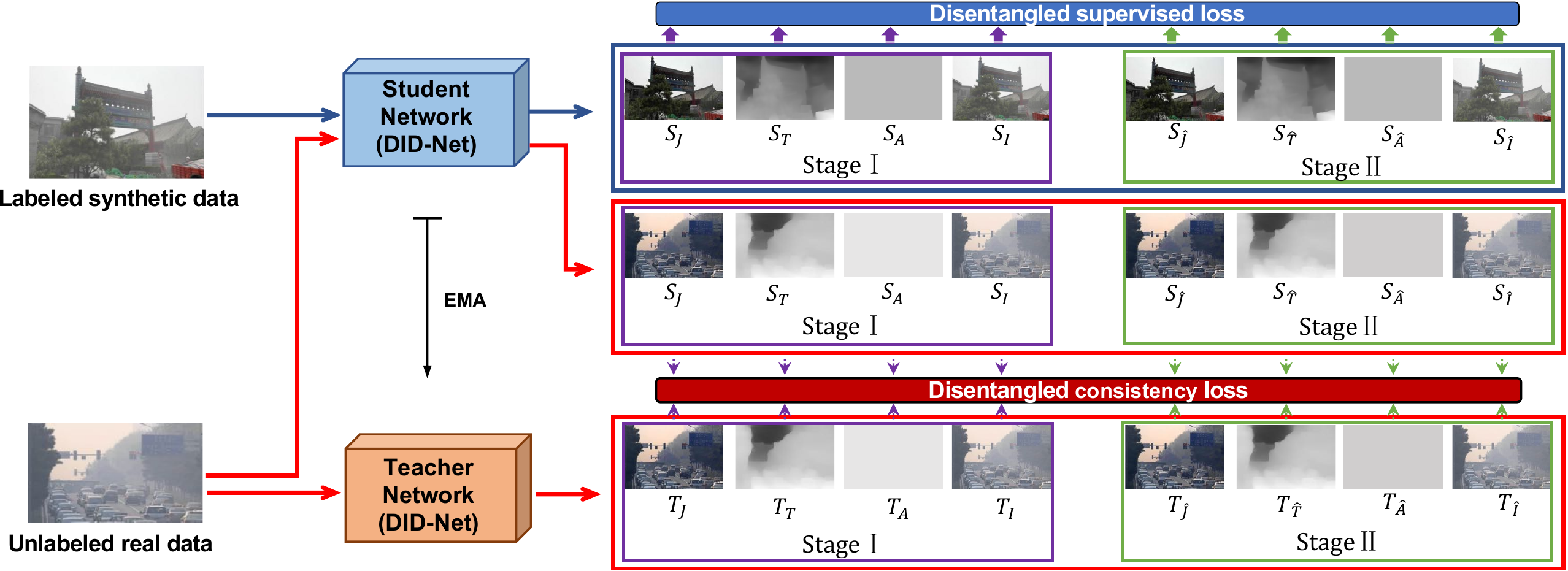}
	\end{center}
	\vspace{-3mm}
	\caption{The schematic illustration of the developed DMT-Net.
		We first develop a disentangled image dehazing network (DID-Net; see Figure~\ref{fig:DCNN}) for detecting haze-free maps, transmission maps, and atmospheric maps via a coarse-to-fine strategy, and build reconstructed haze maps. After that, we compute a supervised loss for labeled data and a consistency loss for unlabeled data and add them together to train our image dehazing netowrk. EMA: exponential moving average.}
	\vspace{-3mm}
	\label{fig:pipeline}
\end{figure*}

\subsection{Traditional Dehazing Methods}
Traditional dehazing methods utilized image priors (e.g., dark channel
prior~\cite{he2011single}, color-line priors~\cite{fattal2014dehazing},
and haze-line priors~\cite{berman2016non}) from hazy and latent clean images to compute a transmission map for haze removal. 
Please refer to Zhang et al.~\cite{zhang2018densely} for a comprehensive review. 
By assuming that a linear relationship exists in the minimum channel between the hazy image and the haze-free image, a single image dehazing method is proposed based on linear transformation~\cite{2017Fast}.
Note that these hand-crafted priors from human observations do not always hold in diverse real-world hazy photos. 
Hence, they tend to suffer from undesirable color distortions~\cite{ren2018gated}.

\vspace{-2mm}
\subsection{Deep Learning based Dehazing Methods}

Early works formulated CNNs to estimate a transmission map for recovering the clean image via the physical model in Eq.~\eqref{Eq:physical}.
Ren et al.~\cite{ren2016single} employed a coarse-scale network to predict a holistic transmission map and then used a fine-scale network for a transmission map refinement.
Cai et al.~\cite{cai2016dehazenet} computed a transmission map by developing a DehazeNet equipped with BReLU based feature extraction layers.
However, inaccurate transmission map estimation hinders haze removal quality of these methods.

Later, CNN-based methods directly learned the latent clean image from a single hazy image in an end-to-end manner.
Yang et al.~\cite{yang2018proximal} predicted a clean image by integrating the physical model and image prior into a CNN.
Song et al.~\cite{2018Single} presented a novel ranking convolutional neural network for single image dehazing.
Li et al.~\cite{Li_2018_CVPR} embedded VGG-features~\cite{simonyan2015very} and
an $L_{1}$-regularized gradient prior into conditional generative adversarial network (cGAN)~\cite{isola2017image} for a clean image estimation.
Ren et al.~\cite{ren2018gated} learned confidence maps via an encoder-decoder network from three derived inputs and fused confidence maps for generating a final dehazed result.
Zheng et al.~\cite{Zheng2021} devised a multi-guide bilateral learning for reaching a real-time dehazing of 4K images.
However, the disjoint optimization on these deep models failed to capture the inherent relations among the transmission map, the atmospheric light, and the dehazed result, thereby degrading the overall dehazing performance.

To alleviate this issue, Zhang et al.~\cite{zhang2018densely} 
employed two networks to estimate the transmission map and the atmospheric light separately, and computed the haze-free image according to the physical haze model (see Eq.~\eqref{Eq:physical}), which are all integrated into an end-to-end dehazing network (DCPDN).
Deng et al.~\cite{deng2019deep} attentively fused multiple mathematical haze separation models for image dehazing.
Qu et al.~\cite{qu2019enhanced} presented a dehazing GAN with a multi-resolution generator module, the enhancer module, and a multi-scale discriminator module.
Liu et al.~\cite{liu2019griddehazenet} devised a CNN with a pre-processing module, an attention-based multi-scale backbone module and a post-processing module.
Deng et al.~\cite{deng2020hardgan} stacked haze-aware representation distillation (HARD) modules with normalization layers into a GAN to attentively fuse global atmospheric brightness and local spatial structures.
Dong et al.~\cite{dong2020physics} explicitly utilized the physical model into a encoder-decoder network.
Although improving the overall scene visibility, these methods are trained on synthesized images in a supervised learning manner, suffering from limited capability to generalize well to real-world hazy images. 

Shao et al.~\cite{shao2020domain} formulated a domain adaptation network with two image translation modules between synthesized and real hazy images and two image dehazing modules to alleviate the domain shift problem.
However, the image dehazing modules of Shao et al.~\cite{shao2020domain} learned CNN features from input hazy image to predict only one factor (i.e., the latent haze-free image), thereby hindering the dehazing performance.
Although Li et al.~\cite{li2020zero} also leveraged the layer separation mechanism based on the physical model, this work mainly addressed the image dehazing in an unsupervised and zero-shot manner.
To alleviate this issue, we develop a disentangled image dehazing network to learn disentangled feature presentations and leverage unlabeled data for improving dehazing performance.

%% file: method.tex

\section{Our Approach}
\label{sec:ourmodel}

Figure~\ref{fig:pipeline} shows the architecture of our DMT-Net, which learns disentangled representations and leverages unlabeled real-world hazy images for image dehazing.
Specifically, we first develop a disentangled image dehazing network (DID-Net) to disentangle features at each scale into three components for jointly computing a dehazed map, a transmission map, and an atmospheric map via a coarse-to-fine mechanism (see predictions at Stage I and Stage II in Figure~\ref{fig:pipeline}).
Moreover, DID-Net reconstructs hazy images from the estimated maps of the three components, and then computes two reconstruction losses between the input hazy image and two reconstructed ones to make them similar.
Then, our DID-Net is assigned to both the student network and the teacher network.
During the training, the labeled data is fed into the student network,
and a supervised loss is computed by adding prediction losses of three component maps and the reconstruction losses.
Then, for unlabeled data, we produce one auxiliary image from the input hazy image and feed it into the student network and the teacher network, respectively.
A disentangled consistency loss is computed on the two groups of three component maps and a reconstructed hazy map.
In the testing procedure, we only utilize the student network to generate the dehazing result of an input image.

\subsection{Disentangled Image Dehazing Network} \label{subsec:DMCNN}

\begin{figure*} [!t]
	\centering
	\includegraphics[width=\linewidth]{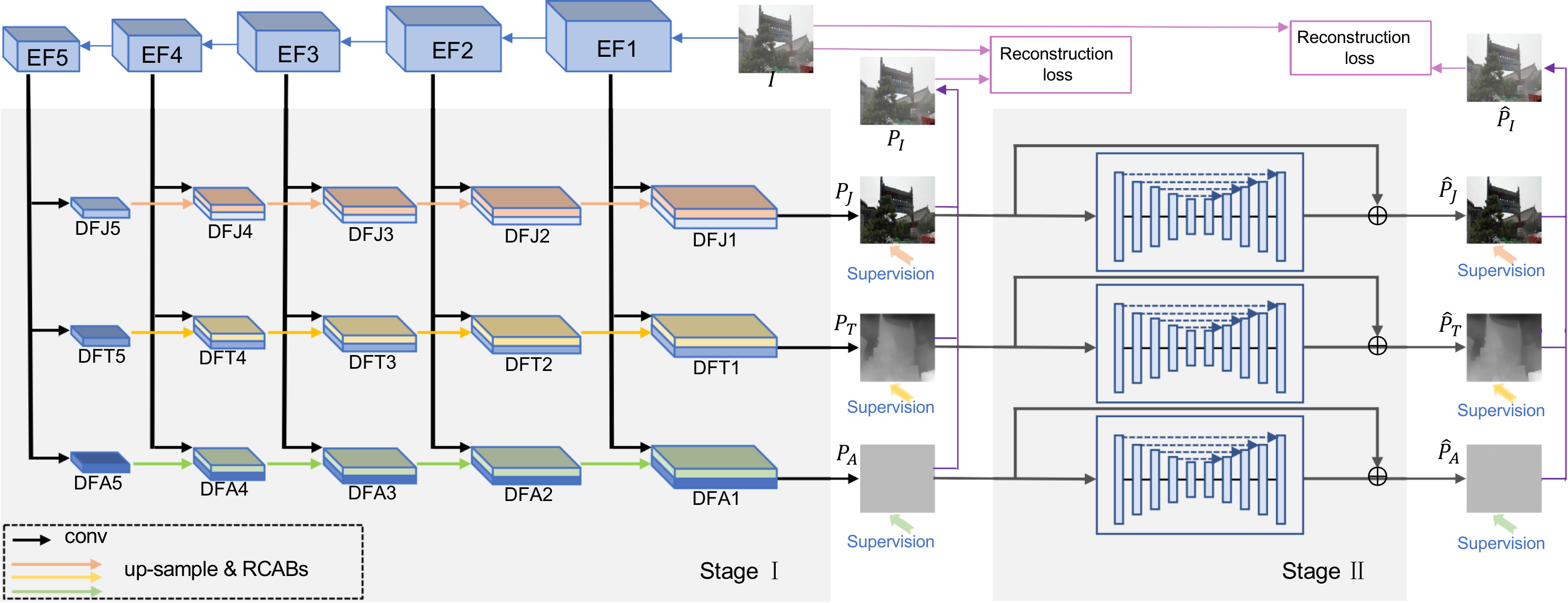} 
    \vspace{-3mm}
	\caption{The schematic illustration of our DID-Net. Given an input hazy image $I$, we first extract a set of feature maps with different spatial resolutions, and then disentangle these  features into three components: dehazing-distilled features for estimating a haze-free image ($J$), transmission-distilled features for estimating a transmission map ($T$), and light-distilled features for estimating an atmospheric light map ($A$). We then devise a coarse-to-fine strategy to predict $J$, $T$, and $A$. The coarse predictions ($P_J$, $P_T$, and $P_A$) are obtained via iteratively merging disentangled features, while refinement predictions ($\hat{P}_J$, $\hat{P}_T$, and $\hat{P}_A$) are produced by feeding coarse results into U-Net structures. Moreover, we reconstruct two hazy images ($P_I$ and $\hat{P}_I$) from the physical haze model with coarse/refined predictions, and compute reconstruction losses against the input hazy image $I$. }
	\label{fig:DCNN}
    \vspace{-3mm}
\end{figure*}

According to the physical model of a haze process (see Eq.~\eqref{Eq:physical}), CNN features learned from an input hazy image encode information of three haze components, i.e. the haze-free map, the transmission map, and the atmospheric map.
In this work, we propose a disentangled image dehazing network (DID-Net) by harnessing a disentangled feature learning strategy~\cite{liu2018exploring} to separate each feature into three components: a latent-distilled feature for predicting the
haze-free map $J$, a transmission-distilled feature for predicting the transmission $T$, and a light-distilled feature for predicting the atmospheric-light map $A$, respectively; see Figure~\ref{fig:DCNN}.
By doing so, the proposed DID-Net is capable of simultaneously extracting features for all the three components, and hence providing comprehensive information for dehazing. 
Note that, before our work, two recently published papers~\cite{zhang2018densely,dong2020physics} also propose to jointly estimate the three components, but they utilize different networks for component estimation and totally rely on synthesized data, resulting in performance degradation on real-world photos. 

Figure~\ref{fig:DCNN} shows the schematic illustration of the proposed DID-Net.
Specifically, given an input hazy image $I$, we first extract a set of feature maps with different spatial resolutions, and these feature maps are denoted as
$\textit{EF}_i$ $(1\leq i \leq 5)$.
We decompose each into latent-distilled disentangled features $\textit{DFJ}_i$, transmission-distilled disentangled features $\textit{DFT}_i$, and light-distilled disentangled features $\textit{DFA}_i$.
After that, we devise a coarse-to-fine mechanism to estimate $J$, $T$, and $A$ based on these disentangled features.
To do so, we first integrate disentangled features at different scales to generate coarse predictions for the three components (denoted as $P_{J}$, $P_{T}$ and $P_{A}$).
Then, a hazy image $P_{I}$ can be figured out as follows:
\begin{equation}  \label{Eq:P_I}
    P_I(z)=P_J(z)\cdot P_T(z)+P_A(z)\cdot (1-P_T(z)) \,
\end{equation}
where $z$ is the $z$-th pixel. 

We then feed the three coarse predictions, $P_{J}$, $P_{T}$ and $P_{A}$, into three independent U-Net based residual blocks to generate three corresponding refined predictions.  
The three refined predictions are denoted as $\hat{P}_J$, $\hat{P}_T$, and $\hat{P}_A$, respectively, based on which another hazy image $\hat{P}_I$ is computed from $\hat{P}_J$, $\hat{P}_T$, and $\hat{P}_A$:
\begin{equation}  \label{Eq:P_I2}
    \hat{P}_I(z)=\hat{P}_J(z)\cdot \hat{P}_T(z)+\hat{P}_A(z)\cdot (1-\hat{P}_T(z)) \ .
\end{equation}

Once reconstructing two hazy images $P_I$ and $\hat{P}_I$ from the coarse and refined predictions, our DID-Net computes a reconstruction loss ($\mathcal{L}_{rec}$) between the input image $I$ and $P_I$ and $\hat{P}_I$. The $\mathcal{L}_{rec}$ is defined as:
\begin{equation}  \label{Eq:reconstruction_loss}
    \mathcal{L}_{rec} = |I-P_I|_{L_1} + |I-\hat{P}_I|_{L_1} \ ,
\end{equation}
where $|\cdot|_{L_1}$ is the $L_1$ loss function.

\noindent
\textbf{How to generate coarse predictions.} \ As shown in Figure~\ref{fig:DCNN}, our DID-Net devises three independent branches to progressively aggregate disentangled features from deep layers to shallow layers for generating coarse predictions $P_J$, $P_T$, and $P_A$.
Here, we take the branch for generating the haze-free image prediction as an example to describe the workflow.
This branch aggregates $\textit{DFJ}_i$ $(1\leq i \leq 5)$ for predicting the haze-free prediction, where the key operation is to merge features at adjacent layers.
When fusing two adjacent features ($\textit{DFJ}_i$ and $\textit{DFJ}_{i-1}$, $2 \leq i \leq 5$), we up-sample the low-resolution feature ${\textit{DFJ}}_{i}$ to the same spatial resolution with the high-resolution feature $\textit{DFJ}_{i-1}$, enhance upsampled features by feeding them into a series of residual channel attention blocks (RCABs)~\cite{zhang2018image}, and apply a $1$$\times$$1$ convolutional layer on the concatenation of the RCAB-enhanced feature and the high-resolution feature $\textit{DFJ}_{i-1}$ to output a merged feature map, which is denoted as $H_{i-1}$.
The $H_{i-1}$ can be computed by:
\begin{equation}
    \begin{aligned}\label{Eq:merge-operation}
    H_{i-1} &= \mathcal{M}(\textit{DFJ}_i, \textit{DFJ}_{i-1}) \\
     &=  conv(concate(\Phi_{\textit{RCAB}}(\textit{DFJ}_i), \textit{DFJ}_{i-1})) \ ,
    \end{aligned}
\end{equation}
where $\mathcal{M}(\cdot)$ denotes the operation of merging two features;
the $conv$ is a $1$$\times$$1$ convolutional layer; the $\Phi_{\textit{RCAB}}$ is the refinement block consisting of a number of RCABs.
Then, similarly, we merge $H_{i-1}$ with features $\textit{DFJ}$ at the next CNN layer until reaching features $\textit{DFJ}$ (with the largest spatial resolution) at the first CNN layer, and finally pass the resultant features to a $1$$\times$$1$ convolutional layer for predicting $P_J$:
\begin{equation}
    \begin{aligned}\label{Eq:P-J}
    P_J &= conv(\mathcal{M}(\textit{DFJ}_1, H_2)) \ , \\
    H_2 &=  \mathcal{M}(\textit{DFJ}_2, H_3) \ , \\
    H_3 &=  \mathcal{M}(\textit{DFJ}_3, H_4) \ , \\
    H_4 &=  \mathcal{M}(\textit{DFJ}_4, \textit{DFJ}_5) \ , \\
    \end{aligned}
\end{equation}
where $\mathcal{M}$ is the feature operation of Eq.~\eqref{Eq:merge-operation}.
We conduct the similar operations to incrementaly aggregate transmission-distilled  features and light-distilled features.
Note that the numbers of RCABs in merging features are different. 
In our experiments, we empirically use $20$ RCABs to merge adjacent features for computing $P_J$ and $P_T$, and $2$ RCABs to merge features for computing $P_A$, since $P_A$ is a global parameter and simpler than the other two components. 

%
%

\noindent
\textbf{How to generate fine predictions.} 
We further pass these coarse predictions, $P_J$, $P_T$, and $P_A$, to a U-Net residual block to produce their refined results.
For example, given the coarse haze-free map prediction $P_J$, we pass it to a U-Net~\cite{ronneberger2015u} with $5$ convolutional layers, to produce an intermediate image $\mathcal{U}(P_J)$, which is then added with $P_J$ to obtain a refinement (denoted as $\hat{P}_J$) of $P_J$,
\begin{equation}  \label{Eq:P_J-refinement}
    \hat{P}_J = P_J + \mathcal{U}(P_J) \ .
\end{equation}
Similarly, we compute a refinement $\hat{P}_T$ of $P_T$, and a refinement $\hat{P}_A$ of $P_A$ as follows:
\begin{equation}
    \begin{aligned}\label{Eq:refinement-P_T-P-A}
    \hat{P}_T & = P_T + \mathcal{U}(P_T) \ , \\
    \hat{P}_A & = P_A + \mathcal{U}(P_A) \ ,
    \end{aligned}
\end{equation}
where $\mathcal{U}(P_T)$ and $\mathcal{U}(P_T)$ are the U-Net structure on $P_T$ and $T_A$. Note that $\mathcal{U}(P_J)$, $\mathcal{U}(P_T)$, and $\mathcal{U}(P_A)$ have the same encoder-decoder structures, but do not share network parameters.

\subsection{Supervised Loss on Labeled data} \label{subsec:supervised loss}


Note that a synthesized hazy image (labeled image) is usually generated by passing a given clean image, a given transmission image, and a given atmospheric image to the physically-based model introduced by Eq.~\eqref{Eq:physical}, which can be naturally taken as the ground truths.
Based on the ground truths, we first compute a disentangled multi-task supervised loss (denoted as $\mathcal{L}_{dst}(x)$) for a labeled hazy image ($x$) by adding the supervised losses of the clean image prediction ($\mathcal{L}^s_J$), transmission image prediction ($\mathcal{L}^s_T$), and atmospheric image prediction ($\mathcal{L}^s_A$), i.e.
\begin{equation}\label{Eq:dst_supervised_loss}
    \mathcal{L}_{dst}(x)= \mathcal{L}^s_J + \alpha_1 \mathcal{L}^s_T + \alpha_2  \mathcal{L}^s_A \ ,
\end{equation}
where
\begin{equation}\label{Eq:dst_supervised_loss_2}
  \begin{aligned}
    & \mathcal{L}^s_J = |G_J-P_J|_{L_1} + |G_J-\hat{P}_J|_{L_1} \ , \\
    & \mathcal{L}^s_T = |G_T-P_T|_{L_1} + |G_T-\hat{P}_T|_{L_1} \ , \\
    & \mathcal{L}^s_A = |G_A-P_A|_{L_1} + |G_A-\hat{P}_A|_{L_1} \ . \\
  \end{aligned}
\end{equation}
%
Here, $G_J$, $G_T$ and $G_A$ represent the ground truths of the clean image, the transmission image and the atmospheric image, respectively.
We empirically set the weights $\alpha_1$$=$$0.3$ and $\alpha_2$$=$$0.1$ in the network training.
By adding $\mathcal{L}_{dst}$ with the reconstruction loss of Eq.~\eqref{Eq:reconstruction_loss}, we compute the supervised loss of labeled data as follows:
\begin{equation}\label{Eq:total_supervised_loss}
    \mathcal{L}^s(x)= \mathcal{L}_{dst}(x) + \alpha_3 \mathcal{L}_{rec} \ ,
\end{equation}
where $\alpha_3$$=$$0.1$ in our experiment.

\subsection{Consistency Loss on Unlabeled Data} \label{subsec:consistency-loss}

For the unlabeled real-world data, we pass it into the student network to obtain eight results, which are two clean images (denoted as $S_J$ and $S_{\hat{J}}$), two transmission images ($S_T$ and $S_{\hat{T}}$), and two atmospheric images ($S_A$ and $S_{\hat{A}}$), and two reconstructed hazy images ($S_I$ and $S_{\hat{I}}$).
Meanwhile, by first adding a Gaussian noise into the real-world hazy image and feeding the noisy image into the teacher network, we can generate another two clean images ($T_J$ and $T_{\hat{J}}$), two transmission images ($T_T$ and $T_{\hat{T}}$), two atmospheric images ($T_A$ and $T_{\hat{A}}$), and two reconstructed hazy images ($T_I$ and $T_{\hat{I}}$).
We then enforce the predictions of eight prediction results from the student network and the teacher network to be consistent, resulting in a disentangled multi-task consistency loss ($\mathcal{L}^c$).
Mathematically, $\mathcal{L}^c$ for an unlabeled image (denoted as $y$) is
\begin{equation}\label{Eq:total_consistency_loss}
    \mathcal{L}^c(y) = \mathcal{L}^c_J + \alpha_4 \mathcal{L}^c_T + \alpha_5 \mathcal{L}^c_A + \alpha_6 \mathcal{L}^c_{rec} \ ,
\end{equation}
where
\begin{equation}\label{Eq:total_consistency_loss_2}
  \begin{aligned}
    & \mathcal{L}^c_J = |S_J-T_J|_{L_1} + |S_{\hat{J}}-T_{\hat{J}}|_{L_1}  \ , \\
    & \mathcal{L}^c_T = |S_T-T_T|_{L_1} + |S_{\hat{T}}-T_{\hat{T}}|_{L_1}  \ , \\
    & \mathcal{L}^c_A = |S_A-T_A|_{L_1} + |S_{\hat{A}}-T_{\hat{A}}|_{L_1}  \ , \\
    & \mathcal{L}^c_{rec} = |S_I-T_I|_{L_1} + |S_{\hat{I}}-T_{\hat{I}}|_{L_1}  \ , \\
  \end{aligned}
\end{equation}
$\mathcal{L}^c_J$, $\mathcal{L}^c_T$, $\mathcal{L}^c_A$, and $\mathcal{L}^c_{rec}$ denote the consistency loss of the clean image estimation, the transmission image estimation, the atmospheric image estimation, and the reconstructed hazy image, respectively. For simplicity, we set
$\alpha_4=\alpha_1=0.3$, $\alpha_5=\alpha_2=0.1$, and $\alpha_6=\alpha_3=$$0.1$.

\subsection{Our Network} \label{subsec:our-network}

As a semi-supervised framework, our method fuses labeled synthesized images and unlabeled real-world images for training.
The total loss of our network is
\begin{equation}\label{Equ:total_loss}
 \mathcal{L}_{total} = \sum_{x \in \mathbb{L}} \mathcal{L}^s(x) + \mu \sum_{y \in \mathbb{U}} \mathcal{L}^c(y) \ ,
\end{equation}
where $\mathbb{L}$ and $\mathbb{U}$ denote the labeled dataset and the unlabeled dataset.
$\mathcal{L}^s(x)$ represents the supervised loss (see Eq.~\eqref{Eq:total_supervised_loss}) for a labeled hazy image x of $\mathbb{L}$.
$\mathcal{L}^c(y)$ is the consistency loss (see Eq.~\eqref{Eq:total_consistency_loss}) for a unlabeled hazy image of $\mathbb{U}$.
We follow~\cite{chen2020multi} to apply a time dependent Gaussian warming up function to compute the weight $\mu$: $\mu(t) = \mu_{max} e^{(-5{(1-t/t_{max})}^2)}$, where $t$ denotes the current training iteration and $t_{max}$ is the maximum training iteration. In our experiments, we empirically set $\mu_{max}$$=$$1$.
We minimize $\mathcal{L}_{total}$ to train the student network, and the parameters of the teacher network are updated via the exponential moving average (EMA) strategy with a EMA deay of $0.99$; please refer to ~\cite{laine2016temporal,tarvainen2017mean,chen2020multi} for details.

\subsection{Our unlabeled data} Note that \cite{shao2020domain} provides an unlabeled dataset with $1,000$ real-world hazy images to train a domain adaption network for haze removal.
To conduct fair comparisons, we use the same real-world dataset of~\cite{shao2020domain} as the unlabeled data of our network.

\subsection{Training and Testing Strategies}
\label{subsec:training-testing-strategy}

\textbf{Training parameters.}\
To accelerate the training procedure and reduce the overfitting risk, we initialize the parameters of DID-Net (student network) by ResNeXt~\cite{xie2017aggregated}, which has been well-trained for the image classification task on the ImageNet. Other parameters in the DID-Net are initialized as random values.
We implement our framework in PyTorch and utilize ADAM optimizer to train the network.
The learning rate is adjusted by a poly strategy~\cite{liu2015parsenet} with the initial learning rate of $0.0001$ and the power of $0.9$.
We randomly crop all the labeled and unlabeled images to $240$$\times$$240$ for the training on two GTX 2080Ti GPU, and augment the training set by random horizontal flipping.
We use the mini-batch size of $16$, which means the usage of $8$ labeled images and $8$ unlabeled data images in each training epoch.

\textbf{Inference.} \
In the testing stage, we feed the input image into the student network and utilize the predicted dehazed map of the student network as the final output of our network.

%% file: ex.tex


\section{Experimental Results}
\label{sec:results}

\begin{table}[!t]
\begin{center}
  \caption{Quantitative comparisons between our network and compared methods on three synthetic dehazing datasets.}
 \vspace{-1.5mm}
  \resizebox{1\linewidth}{!}{%
  \label{table:O-HAZE}
    \begin{tabular}{c|c|c|c|c|c|c|c}
        & &
        \multicolumn{2}{c|}{Haze4K} &
        \multicolumn{2}{c|}{SOTS~\cite{ren2018gated}} &
        \multicolumn{2}{c}{HazeRD~\cite{zhang2017hazerd}}
        \\
        \hline
        method &Year &PSNR & SSIM & PSNR & SSIM & PSNR & SSIM 
        \\
        \hline
        \hline
        \textbf{Our DMT-Net} &-   & \textbf{28.53} & \textbf{0.96}  & \textbf{29.42} &\textbf{0.97}
        & \textbf{18.55} &\textbf{0.85}
        \\ 
        Our DID-Net &- &27.81   &0.95  &28.30 &0.95 &18.07 &0.84
        \\  
        \hline \hline
        DA~\cite{shao2020domain} &2020  &24.03 &0.90 &27.76 &0.93 & 18.07 &0.63   \\
        FFA-Net~\cite{2020FFA} &2020  &26.97 &0.95 &26.88 &0.95 &17.56 &0.80  \\
        MSBDN~\cite{MSBDN-DFF} &2020  &22.99 &0.85 &24.15 &0.86 & 16.87 &0.75  \\
        DM$^2$F-Net~\cite{deng2019deep} &2019 &24.61 &0.92 &23.87 &0.91 &15.88 &0.74  \\
        GDN~\cite{liu2019griddehazenet} &2019  &23.29 &0.93 &26.05 &0.95 &15.92 &0.77  \\
        EPDN~\cite{qu2019enhanced} &2019 &21.08 &0.86 &23.82 &0.89 &17.37 &0.56 \\
        \hline
        \hline
        DCPDN~\cite{zhang2018densely} &2018  &23.86 &0.91 &19.39 &0.65 &16.12 &0.34  \\
        GFN~\cite{ren2018gated} &2018   &- &-  &22.30 &0.88 &13.98 &0.37    \\
        AOD-Net~\cite{li2017all} &2017  &17.15 &0.83  &19.06 &0.85  &15.63 &0.45 \\
        DehazeNet~\cite{cai2016dehazenet} &2016  &19.12 &0.84  &21.14 &0.85 &15.54 &0.41   \\
        MSCNN~\cite{ren2016single} &2016  &14.01 &0.51   &17.57 &0.81 &15.57 &0.42   \\
        \hline
        \hline
        NLD~\cite{berman2016non} &2016 &15.27 &0.67  &17.27 &0.75  &16.16 &0.58 \\
        DCP~\cite{he2011single} &2011 &14.01 &0.76  &15.49 &0.64  &14.01 &0.39 \\
        \hline
    \end{tabular}
    }
  \end{center}
  \vspace{-5mm}
\end{table}

\input{result-synthetic-figure.tex}

We compare our dehazing network against 13 state-of-the-art image dehazing methods,
including DCP~\cite{he2011single},
NLD~\cite{berman2016non},
MSCNN~\cite{ren2016single},
DehazeNet~\cite{cai2016dehazenet},
AOD-Net~\cite{li2017all},
GFN~\cite{ren2018gated},
DCPDN~\cite{zhang2018densely},
EPDN~\cite{qu2019enhanced},
GDN~\cite{liu2019griddehazenet},
DM$^2$F-Net~\cite{deng2019deep},
FFA ~\cite{2020FFA},
MSBDN~\cite{MSBDN-DFF}
and DA~\cite{shao2020domain}.
Among the compared methods, DCP and NLD focused on hand-crafted features for haze removal, while others are based on convolutional neural networks (CNNs). We retrain the original (released) implementations of these methods or directly report their results on the public datasets.
Furthermore, we employ two widely-used metrics for quantitative comparisons,
and they are peak signal to noise ratio (PSNR)~\cite{zhu2017non,zhu2020learning} and structural similarity index (SSIM)~\cite{wang2004image,zhu2017joint}. 


\noindent
\textbf{Datasets.} \ We first test each image dehazing method on a public benchmark dataset, i.e., SOTS~\cite{shao2020domain}, which consists of 1,000 testing images.
We follow existing works~\cite{shao2020domain} to set the associate training set with 6,000 synthesized images, which consists of 3,000 from the indoor training set (ITS), and 3,000 from the outdoor training set (OTS) of the RESIDE dataset~\cite{li2018benchmarking}.
Second, HazeRD~\cite{zhang2017hazerd} containing $15$ outdoor images with more realistic haze is introduced for testing.

Apart from SOTS and HazeRD, we also create a synthesized dataset (denoted as Haze4K) with 4,000 hazy images, in which each hazy image has the associate ground truths of a latent clean image, a transmission map, and an atmospheric light map.
To be specific, we collected 1,000 clean images by randomly selecting $500$ indoor images from NYU-Depth~\cite{silberman2012indoor} and $500$ outdoor images from OTS~\cite{li2018benchmarking}.
Among them, $250$ images are randomly selected from both indoor image set ($125$ images) and outdoor image set ($125$ images), to form the test set, and the remaining $750$ images are used for the training set.
After that, for each clean image, we followed~\cite{zhang2018densely} to randomly sample four parameter settings, i.e. atmospheric light conditions $A \in [0.5, 1]$ and scattering coefficients $\beta \in [0.5, 2]$, to generate transmission maps and atmospheric light maps, which are then employed to obtain the corresponding hazy images via the physic model in Eq.~\eqref{Eq:physical}.
Hence, Haze4K has 4,000 hazy images with 3,000 training images and 1,000 testing images.

\input{result-real-figure1}

\input{result-real-figure2}


\vspace{-3mm}
\subsection{Results on Synthetic Images}

We retrain released models of these compared methods on the training set of our Haze4K dataset to obtain their results, while we follow the training setting of DA~\cite{shao2020domain} to produce our results on SOTS and HazeRD for fair comparisons.

Table~\ref{table:O-HAZE} reports PSNR and SSIM scores of different dehazing methods.
In general, CNN-based methods have larger
PNSR and SSIM values than hand-crafted-prior based methods (DCP \& NLD).
Among all the compared methods, FFA-Net has the largest PSNR and SSIM scores (i.e., 26.97 and 0.95) on Haze4K, and the largest SSIM score (0.80) on HazeRD, while DA has the largest PSNR and SSIM values (i.e., 27.76 and 0.93) on SOTS, and the largest SSIM value (18.07) on HazeRD.
Also note that our Haze4K dataset contains more challenging dehazing photos than SOTA, and existing dehazing methods suffer from a degraded PSNR and SSIM performance.
DID-Net, as our sub-network with only labeled data, already outperforms most existing CNN-based methods in terms of PSNR and SSIM metrics, which proves the effectiveness of our disentangled feature learning for haze removal.
Furthermore, our method consistently has the largest PSNR and SSIM scores on Haze4K, SOTS, and HazeRD,
demonstrating that our semi-supervised dehazing network can better recover the underlying clean images for these hazy images.

Figure~\ref{fig:synthetic} visually compares the dehazed results.
In the first and third images, DehazeNet produces an obvious color distortion in the ground regions of the dehazed results.
Although obtaining a better dehazing performance than DehazeNet, the CNN-based methods (e.g., GDN, DM$^2$F-Net, FFA, and MSBDN) tend to darken several areas in their results; see Figures~\ref{fig:synthetic} (c)-(f). DA may produce color distortion especially in the sky regions for the three images. 
In contrast, the dehazed results of our network in Figure~\ref{fig:synthetic} (h) is closest to the latent ground truth images (see Figure~\ref{fig:synthetic} (i)).
To summarize, our dehazed results (DMT-Net) tend to produce higher visual quality and less color distortions, which are also verified by the largest PSNR and SSIM values shown in Figure~\ref{fig:synthetic}.

\vspace{-2mm}
\subsection{Results on Real-world Images}

Figure~\ref{fig:comparisons_real_part1} and Figure~\ref{fig:comparisons_real_part2} visually compare the dehazed maps on real-world hazy photos from the RESIDE dataset~\cite{li2018benchmarking}.
DA suffers from color distortions in almost all the five photos. This is particularly evident in the first and third images of Figure~\ref{fig:comparisons_real_part1}. 
GDN tends to darken several areas; see the first image (the lane area) of Figure~\ref{fig:comparisons_real_part2}.
AOD-Net, DM$^2$F-Net, FFA, and MSBDN remove few fog, and there is still a large amount of fog in the generated images; see blown-up views of Figure~\ref{fig:comparisons_real_part2}.
Our method can more effectively remove haze while producing realistic colors than these compared state-of-the-art methods.

\input{result-AB-figure}

\begin{table}[!t]
\begin{center}
  \caption{Average PSNR and SSIM values in ablation study.}
  \vspace{-3mm}
  \resizebox{0.95\columnwidth}{!}{%
  \label{table:ablation_study}
    \begin{tabular}{c|c|c|c|c}
        &
        \multicolumn{2}{c|}{Haze4K} &
        \multicolumn{2}{c}{SOTS~\cite{ren2018gated}}
        \\
        \hline
        method & PSNR & SSIM & PSNR & SSIM
        \\
        \hline
        \hline
        basic & 25.39 & 0.93  & 27.09 & 0.94
        \\
        \hline
        \hline
        basic+StageI &26.85   &0.94  &27.82 &0.95
        \\
        basic+two-stages &27.81   &0.95  &28.30 &0.95
        \\
        \hline
        \hline
        \textbf{DMT-Net (ours)} & \textbf{28.53} & \textbf{0.96}  & \textbf{29.42} &\textbf{0.97}
        \\
        \hline
    \end{tabular}
    }
  \end{center}
  \vspace{-5mm}
\end{table}

\subsection{Ablation Study}

\noindent
\textbf{Baseline network setting.} \ We perform ablation study experiments to evaluate the effectiveness of major components of our network.
Here, we construct three baseline networks, and list their PSRN and SSIM results on Haze4K and SOTS~\cite{li2019benchmarking,ren2018gated}.
The first baseline (denoted as ``basic'') is constructed by 
removing the two branches of predicting transmission maps and atmospheric maps, removing prediction refinement, and removing unlabeled data.
It means that ``basic'' is equal to progressively merge ${EF}_5$ to ${EF}_1$ on labeled data for predicting a haze-free map $P_J$. 
The second baseline (denoted as ``basic+StageI'') adds the feature disentangling operations into ``basic'', demonstrating that three branches to fuse disentangled features are employed.
Lastly, we construct the third baseline (denoted as ``basic+two-stages'') by adding U-Net refinement blocks to coarse predictions, which equals to train DID-Net (see Fig.~\ref{fig:DCNN}) on labeled data for haze removal.

\noindent
\textbf{Quantitative comparison.} \ Table~\ref{table:ablation_study} summarizes PSNR and SSIM results of our method (DMT-Net) and the constructed three baselines.
From the results, we find that ``basic+StageI'' has larger PSNR and SSIM scores than ``basic", which indicates disentangling CNN features from the input hazy can produce a more accurate dehazed result.
Similarly, ``basic+two-stages'' has a superior PSNR and SSIM performance than ``basic+StageI'', showing that three refinement blocks further improve the dehazing performance.
Lastly, our DMT-Net outperforms ``basic+two-stages'' in terms of PSNR and SSIM metrics. It further demonstrates that the unlabeled data helps our method to obtain better performance.

 \noindent
 \textbf{Visual comparison.} \
As shown in Figure~\ref{fig:result-AB-figure},
``basic+StageI'' tends to produce a relatively clean texture than ``basic'', but both of them tend to modify colors of the building regions. This phenomenon is improved by ``basic+two-stages'', and our method can further generates better textures and visual quality.
In conclusion, our network can effectively remove haze and simultaneously maintain the latent color distributions inside building regions, which is also proved by superior PSRN/SSIM scores.

\vspace{-5mm}
\subsection{More Discussions}

\noindent
\textbf{Hyper-parameter study.} \  As presented in Eq.~\ref{Eq:dst_supervised_loss} and Eq.~\ref{Eq:total_consistency_loss}, our supervised loss on labeled data and unsupervised loss on unlabeled data contain six hyper-parameters to weight different loss functions, and we set them as: $\alpha_4$=$\alpha_1$, $\alpha_5$=$\alpha_2$, and $\alpha_6$=$\alpha_3$.
Table~\ref{table:hyper-parameters} shows the quantitative results of our network and other its modifications. 
We can see that different settings of these hyper-parameters have a certain impact on the dehazed results, and overall they all achieve good results.

\noindent
\textbf{Model complexity analysis.} \ 
The model complexity and inference time of our method are 51.79M/0.127s, worse than the lightweight model AOD-Net (1761/0.004s). We take the task of reducing the model complexity and inference time as one of our future work.

\noindent
\textbf{Results of the teacher model.} \ 
The dehazed PSNR/SSIM of the teacher network are 28.34/0.96, only slightly worse than the student network. 
Following all research works based on the mean-teacher framework, we also utilized the student network to do the inference.

\begin{table}[!t]
\begin{center}
  \caption{Average PSNR and SSIM values of our network on Haze4K under different hyper-parameter settings in Eq.~\ref{Eq:dst_supervised_loss} and Eq.~\ref{Eq:total_consistency_loss}.}
  \vspace{-1.5mm}
  \resizebox{0.8\columnwidth}{!}{%
  \label{table:hyper-parameters}
    \begin{tabular}{c|c|c|c|c|c}
        \hline
        &$\alpha_1(\alpha_4)$ &$\alpha_2(\alpha_5)$ &$\alpha_3(\alpha_6)$ & PSNR & SSIM 
        \\
        \hline
         \hline
        $M_1$ & 0.3 &0.7 &0.1 & 28.70 & 0.97
        \\
        \hline
        $M_2$ &0.3 &0.7 &0.7 &28.40   &0.96
        \\
        \hline
        $M_3$ &0.7 &0.7 &0.1 &28.84   &0.97
        \\
        \hline
        $M_4$ &0.7 &0.1 &0.7 & 28.07 &0.96 
        \\
         \hline
        Ours &0.7 &0.1 &0.7 & 28.53 &0.96 
        \\
        \hline
    \end{tabular}
    }
  \end{center}
  \vspace{-5mm}
\end{table}




%% file: result-synthetic-figure.tex
\begin{figure*}	[t]
	\centering 
	\resizebox{\linewidth}{!}{
		\setlength{\tabcolsep}{1.35mm}
		\begin{tabular}{ccccccccc}
			\Huge \raisebox{0.2\height}{PSNR / SSIM} &
			\Huge \raisebox{0.2\height}{21.80 / 0.70} &
			\Huge \raisebox{0.2\height}{24.63 / 0.94}&
			\Huge \raisebox{0.2\height}{21.99/0.92}&
			\Huge \raisebox{0.2\height}{22.70 / 0.86}&
			\Huge \raisebox{0.2\height}{22.17 / 0.87}&
		    \Huge \raisebox{0.2\height}{27.87 / 0.95}&
			\Huge \raisebox{0.2\height}{28.63 / 0.97}&
			\Huge \raisebox{0.2\height}{$\infty$ / 1}\\
			
			\includegraphics[width=.3\linewidth]{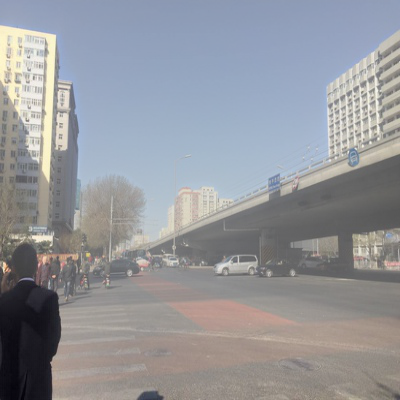}& 
        	\includegraphics[width=.3\linewidth]{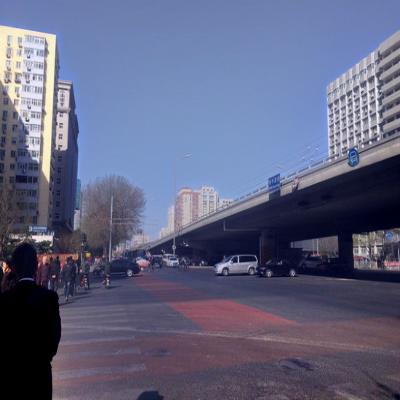}&
			\includegraphics[width=.3\linewidth]{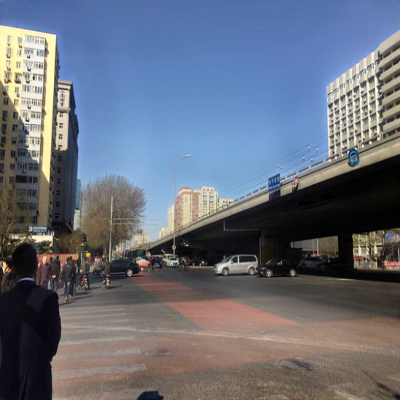}&
			\includegraphics[width=.3\linewidth]{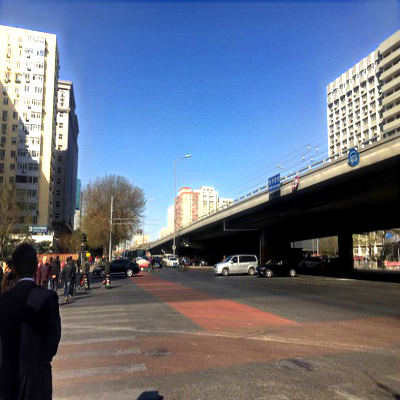}& 
			\includegraphics[width=.3\linewidth]{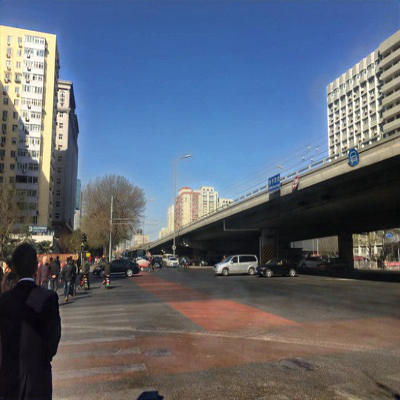} &
			\includegraphics[width=.3\linewidth]{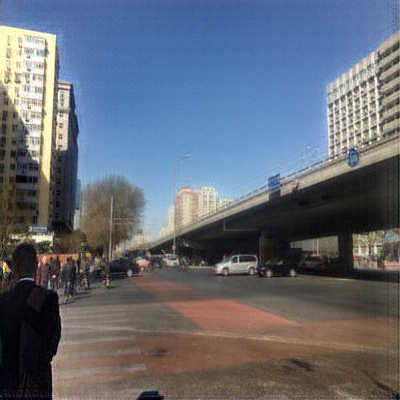}&
			\includegraphics[width=.3\linewidth]{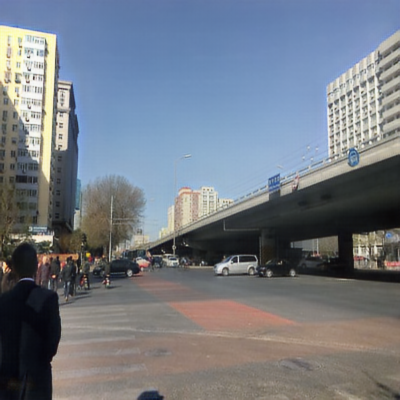}&
			\includegraphics[width=.3\linewidth]{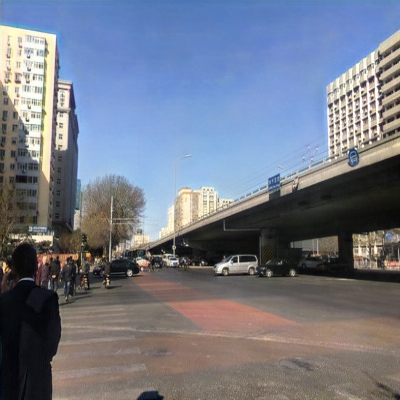}&
			\includegraphics[width=.3\linewidth]{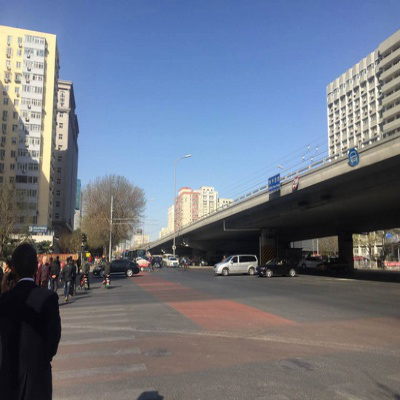}\\

			\Huge\raisebox{0.2\height}{PSNR / SSIM} &
			\Huge\raisebox{0.2\height}{17.76 / 0.83} &
			\Huge\raisebox{0.2\height}{22.92 / 0.96}&
			\Huge\raisebox{0.2\height}{20.29/ 0.86}&
			\Huge\raisebox{0.2\height}{22.34 / 0.94}&
			\Huge\raisebox{0.2\height}{18.92 / 0.84}&
			\Huge\raisebox{0.2\height}{24.99 / 0.93}&
			\Huge\raisebox{0.2\height}{27.69 / 0.96}&
			\Huge\raisebox{0.2\height}{$\infty$ / 1}\\
			
			\includegraphics[width=.3\linewidth]{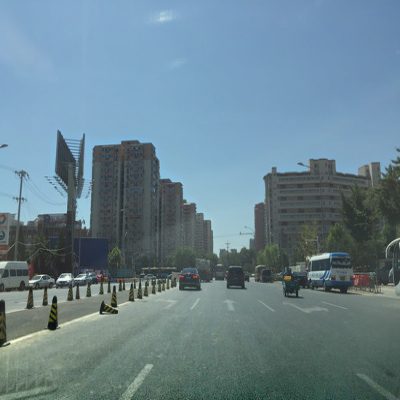}& 
			\includegraphics[width=.3\linewidth]{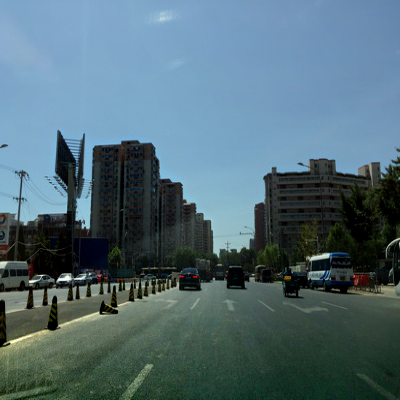}&
			\includegraphics[width=.3\linewidth]{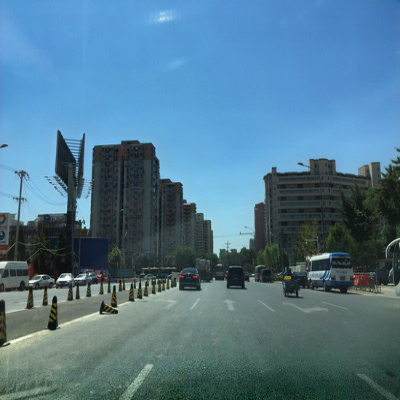}&
			\includegraphics[width=.3\linewidth]{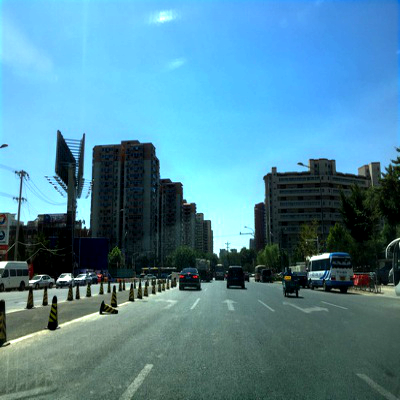}& 
			\includegraphics[width=.3\linewidth]{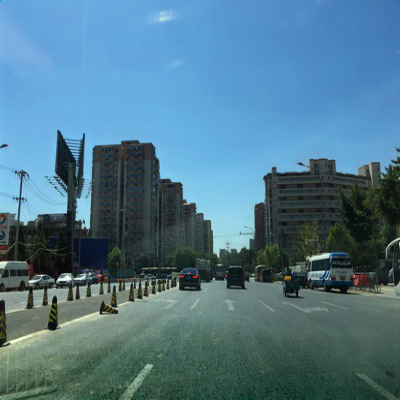} &
			\includegraphics[width=.3\linewidth]{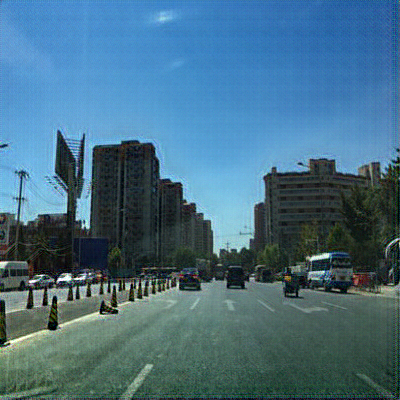}&
			\includegraphics[width=.3\linewidth]{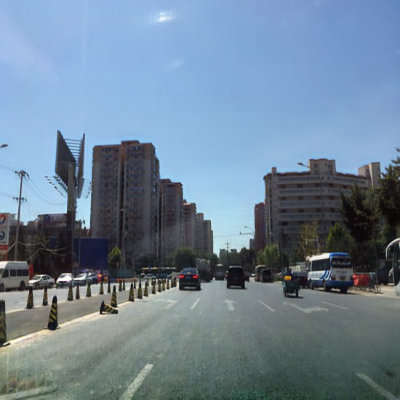}&
			\includegraphics[width=.3\linewidth]{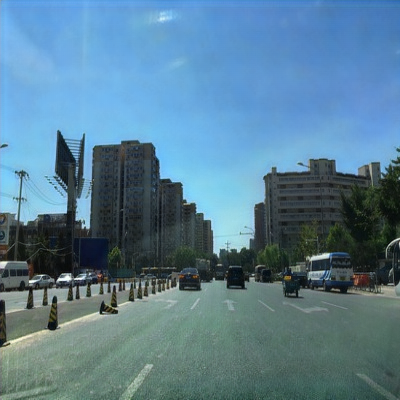}&
			\includegraphics[width=.3\linewidth]{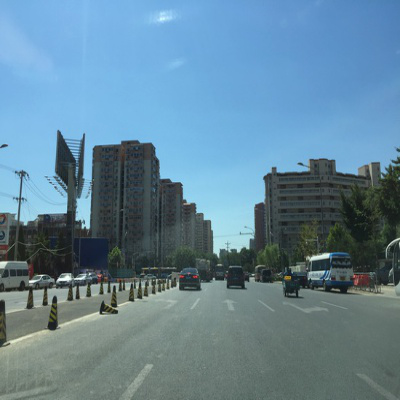}\\
			
			\Huge\raisebox{0.2\height}{PSNR / SSIM} &
			\Huge\raisebox{0.2\height}{10.64 / 0.72} &
			\Huge\raisebox{0.2\height}{18.99 / 0.91}&
			\Huge\raisebox{0.2\height}{18.63/ 0.94}&
			\Huge\raisebox{0.2\height}{21.67 / 0.95}&
			\Huge\raisebox{0.2\height}{17.72 / 0.83}&
			\Huge\raisebox{0.2\height}{17.03 / 0.86}&
			\Huge\raisebox{0.2\height}{27.14 / 0.96}&
			\Huge\raisebox{0.2\height}{$\infty$ / 1}\\
			\includegraphics[width=.3\linewidth]{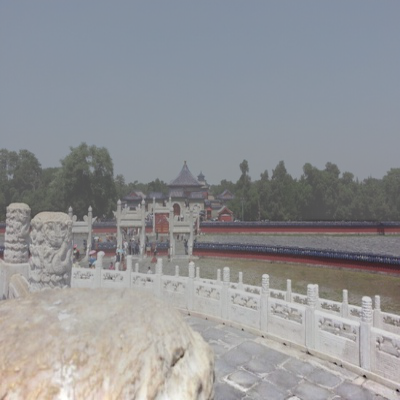}& 
			\includegraphics[width=.3\linewidth]{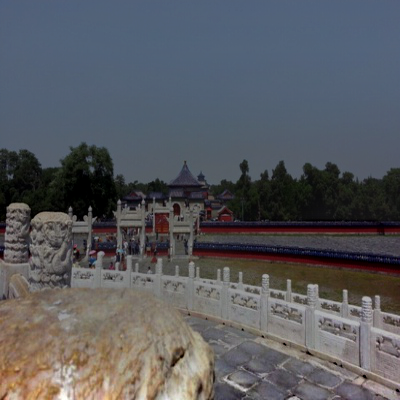}&
			\includegraphics[width=.3\linewidth]{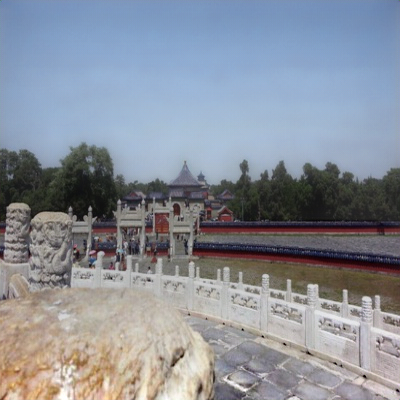}&
			\includegraphics[width=.3\linewidth]{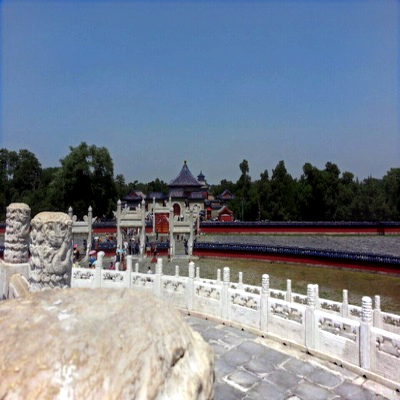}& 
			\includegraphics[width=.3\linewidth]{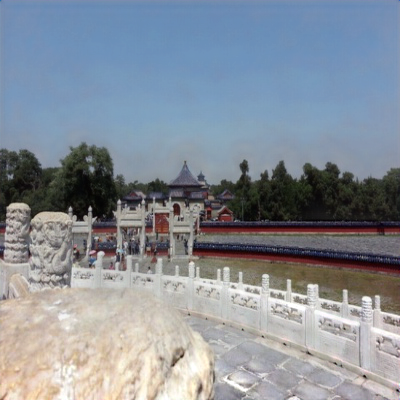} &
			\includegraphics[width=.3\linewidth]{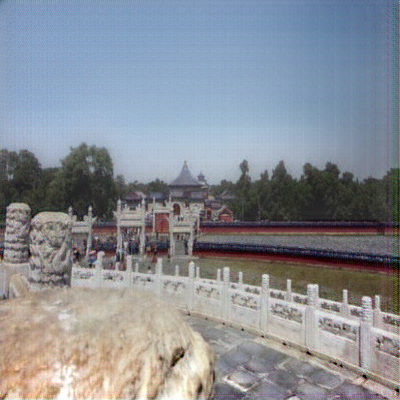}&
			\includegraphics[width=.3\linewidth]{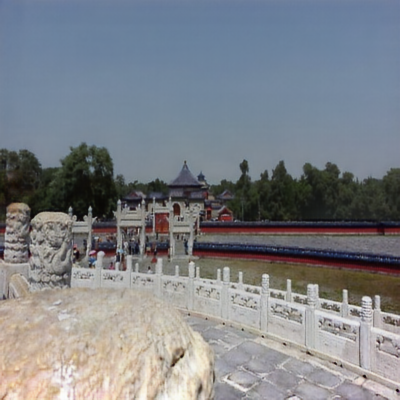}&
			\includegraphics[width=.3\linewidth]{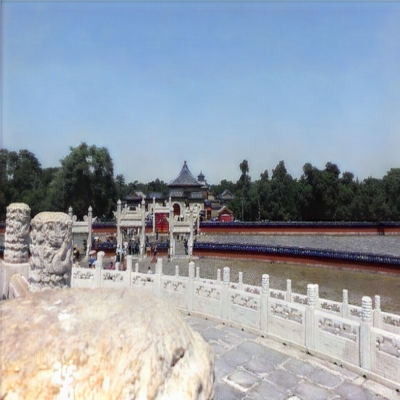}&
			\includegraphics[width=.3\linewidth]{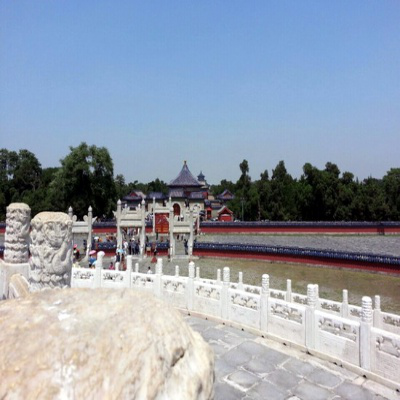}\\
			
			\Huge\raisebox{-0.2\height}{(a) Input Image} &
			\Huge\raisebox{-0.2\height}{(b) DehazeNet} &
			\Huge\raisebox{-0.2\height}{(c) GDN}&
			\Huge\raisebox{-0.2\height}{(d) DM$^2$F-Net }&
			\Huge\raisebox{-0.2\height}{(e) FFA-Net}&
			\Huge\raisebox{-0.2\height}{(f) MSBDN}&
			\Huge\raisebox{-0.2\height}{(g) DA}&
			\Huge\raisebox{-0.2\height}{(h) Our Method}&
			\Huge\raisebox{-0.2\height}{(i) Ground truth}\\
		\end{tabular}}
		\vspace{-3mm}
		\caption{Visual comparisons on dehazed results of various methods on synthetic hazy photos. Please zoom in for a better illustration.
		}
		\label{fig:synthetic}
	\end{figure*}

%% file: result-real-figure1.tex
\begin{figure*}	[t]
	\centering 
	\resizebox{\linewidth}{!}{
		\setlength{\tabcolsep}{1.35mm}
		\begin{tabular}{cccccccc}		
			\includegraphics[width=.3\linewidth]{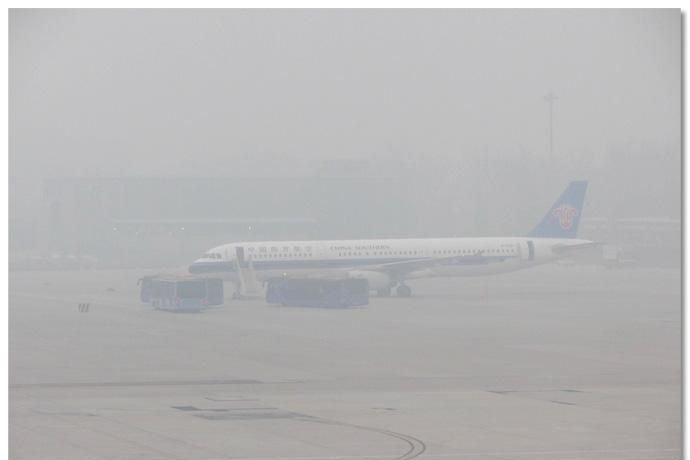}& 
        	\includegraphics[width=.3\linewidth]{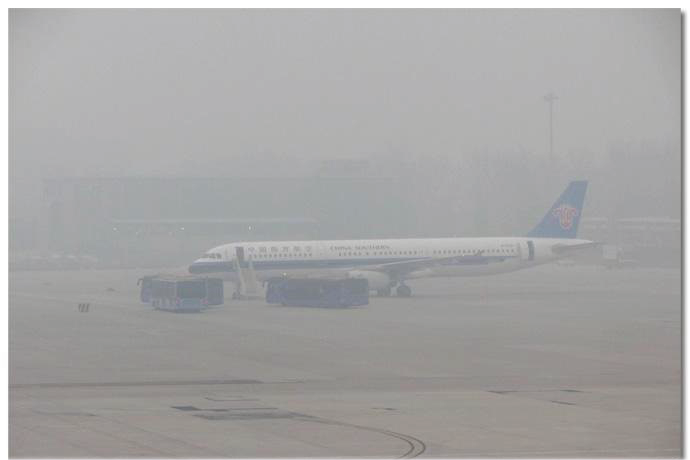}&
			\includegraphics[width=.3\linewidth]{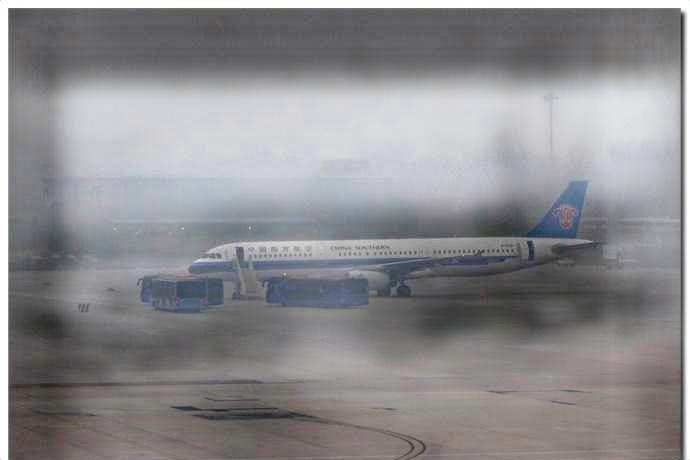}&
			\includegraphics[width=.3\linewidth]{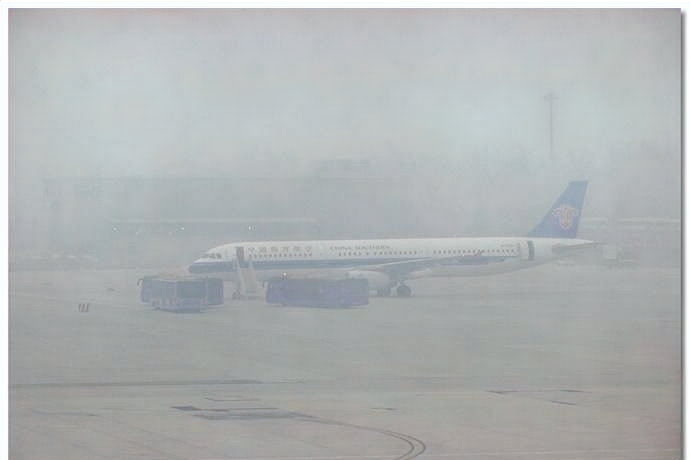}&
			\includegraphics[width=.3\linewidth]{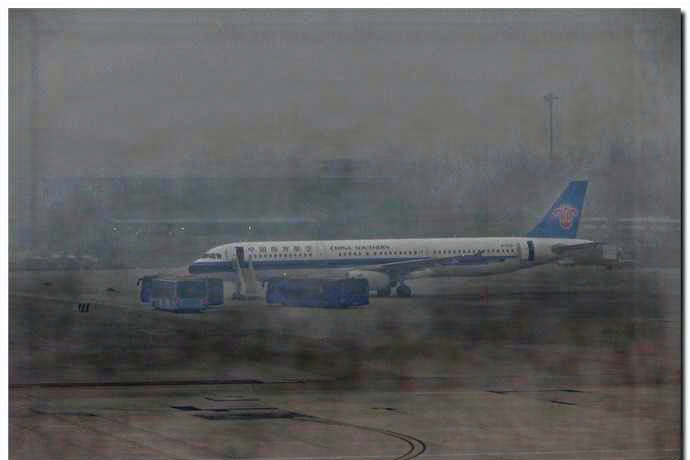} &
			\includegraphics[width=.3\linewidth]{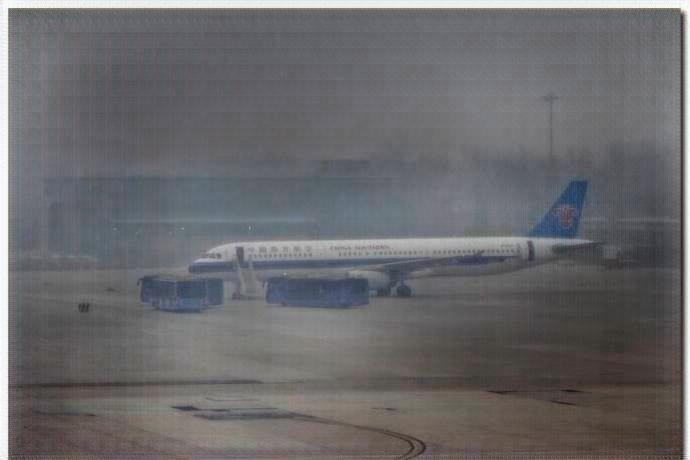}&
			\includegraphics[width=.3\linewidth]{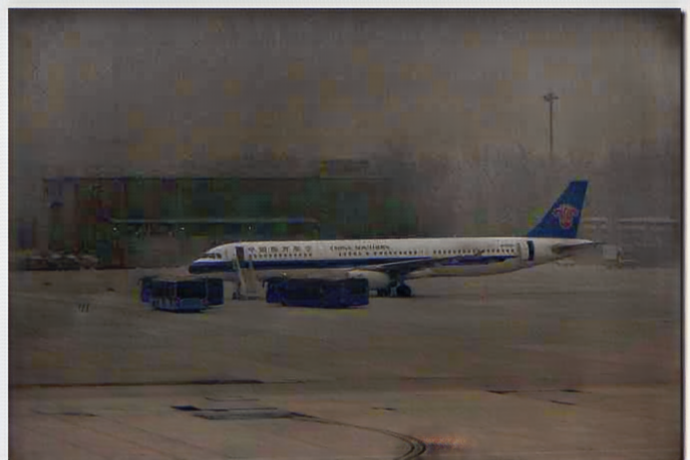}&
			\includegraphics[width=.3\linewidth]{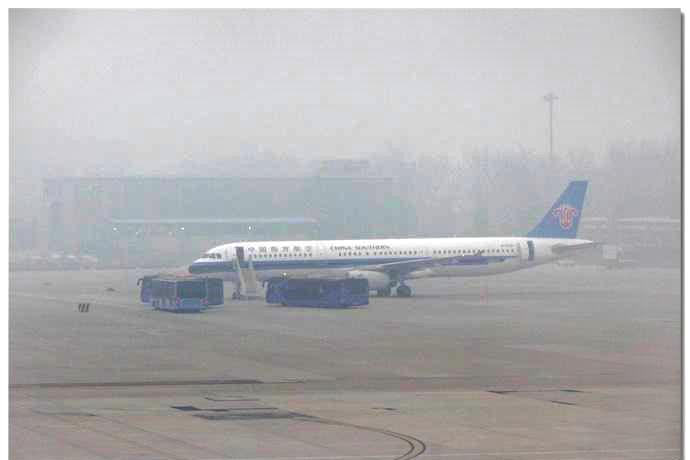}\\

			\includegraphics[width=.3\linewidth]{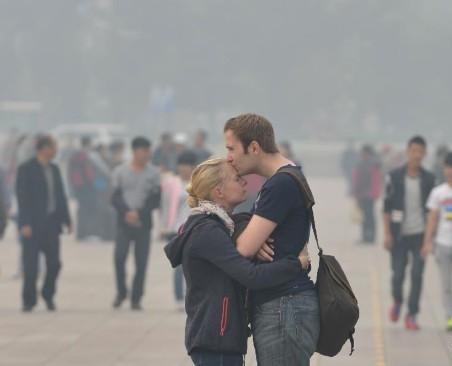}& 
        	\includegraphics[width=.3\linewidth]{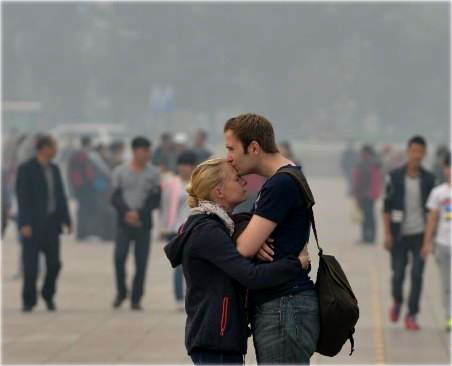}&
			\includegraphics[width=.3\linewidth]{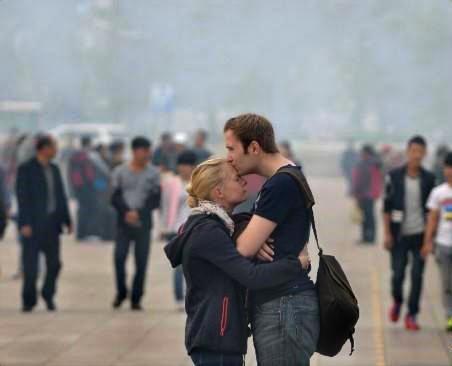}&
			\includegraphics[width=.3\linewidth]{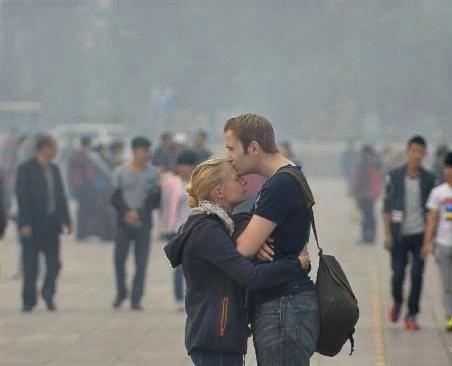}&
			\includegraphics[width=.3\linewidth]{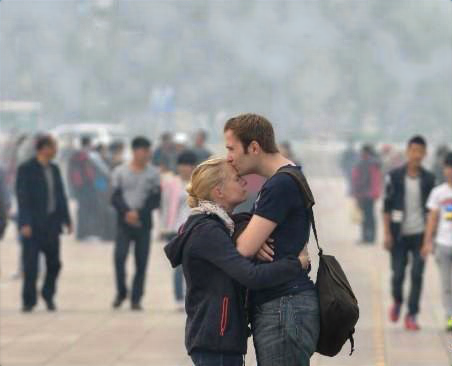} &
			\includegraphics[width=.3\linewidth]{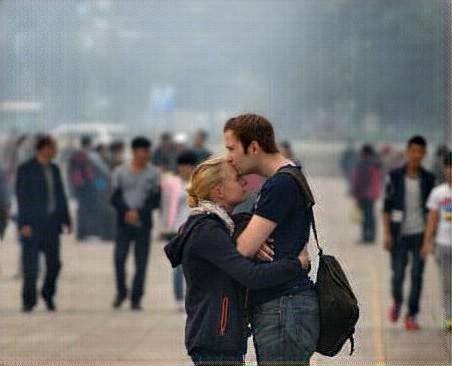}&
			\includegraphics[width=.3\linewidth]{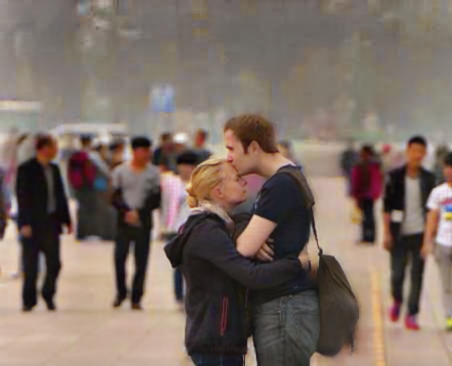}&
			\includegraphics[width=.3\linewidth]{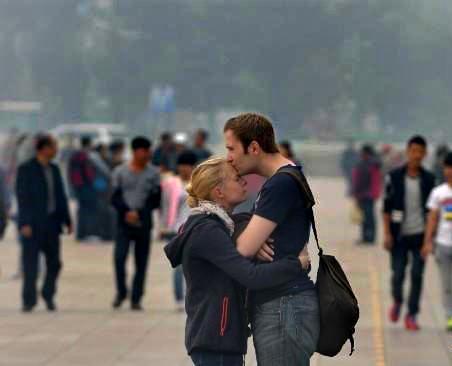}\\

			\includegraphics[width=.3\linewidth]{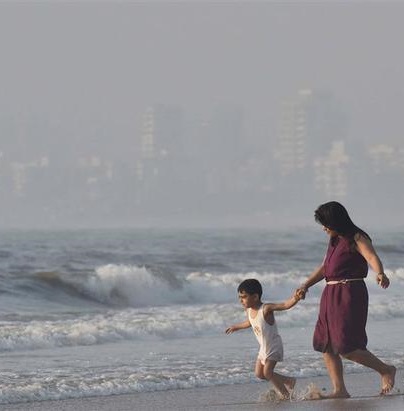}& 
        	\includegraphics[width=.3\linewidth]{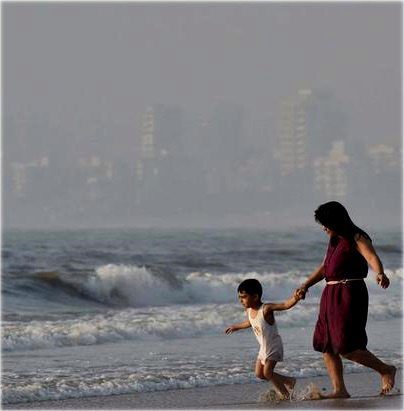}&
			\includegraphics[width=.3\linewidth]{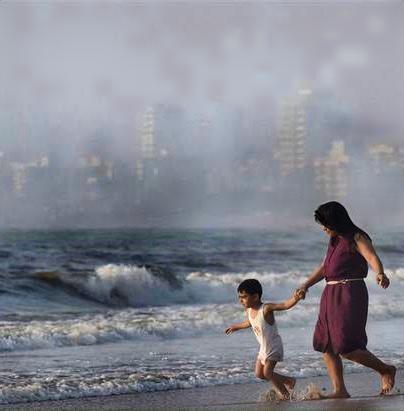}&
			\includegraphics[width=.3\linewidth]{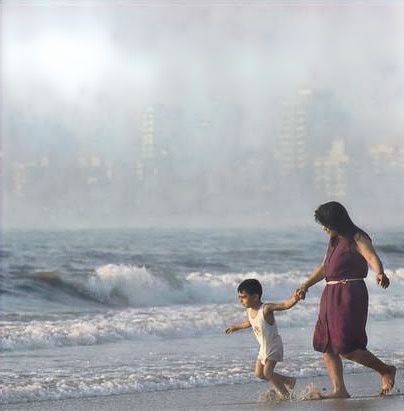}& 
			\includegraphics[width=.3\linewidth]{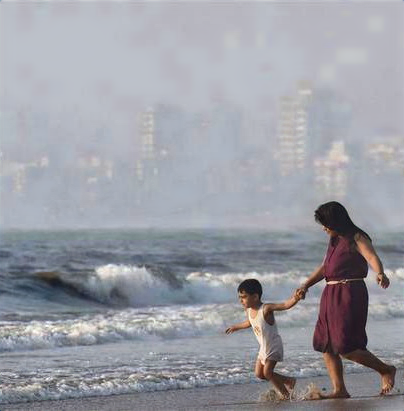} &
			\includegraphics[width=.3\linewidth]{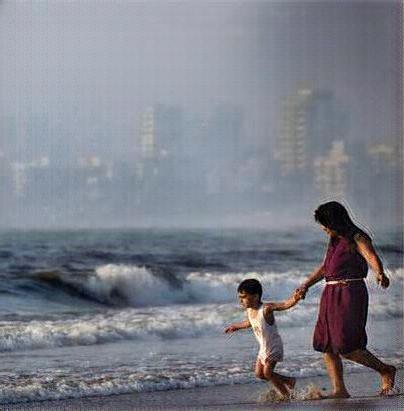}&
			\includegraphics[width=.3\linewidth]{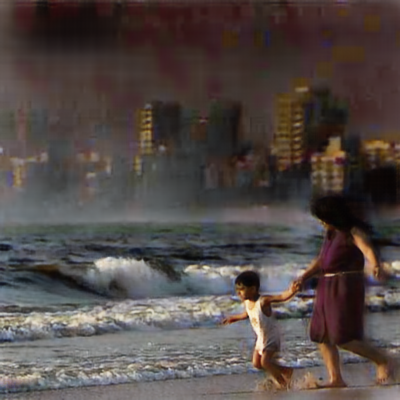}&
			\includegraphics[width=.3\linewidth]{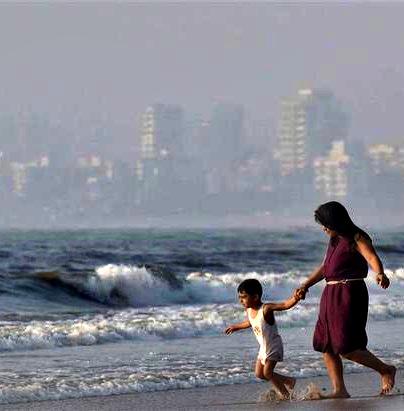}\\

			\Huge\raisebox{-0.2\height}{(a) Input Image} &
			\Huge\raisebox{-0.2\height}{(b) AOD-Net} &
			\Huge\raisebox{-0.2\height}{(c) GDN}&
			\Huge\raisebox{-0.2\height}{(d) DM$^2$F-Net }&
			\Huge\raisebox{-0.2\height}{(e) FFA-Net}&
			\Huge\raisebox{-0.2\height}{(f) MSBDN}&
			\Huge\raisebox{-0.2\height}{(g) DA}&
			\Huge\raisebox{-0.2\height}{(h) Our Method}\\
		\end{tabular}}
		\vspace{-3mm}
		\caption{Visual comparisons on dehazed results produced by our network (h) and SOTA methods (b)-(g) on real-world hazy photos (a).
		}
		\label{fig:comparisons_real_part1}
		\vspace{-3mm}
	\end{figure*}

%% file: result-real-figure2.tex
\begin{figure*}	[t]
	\centering 
	\resizebox{\linewidth}{!}{
		\setlength{\tabcolsep}{1.35mm}
		\begin{tabular}{cccccccc}

			\includegraphics[width=.3\linewidth]{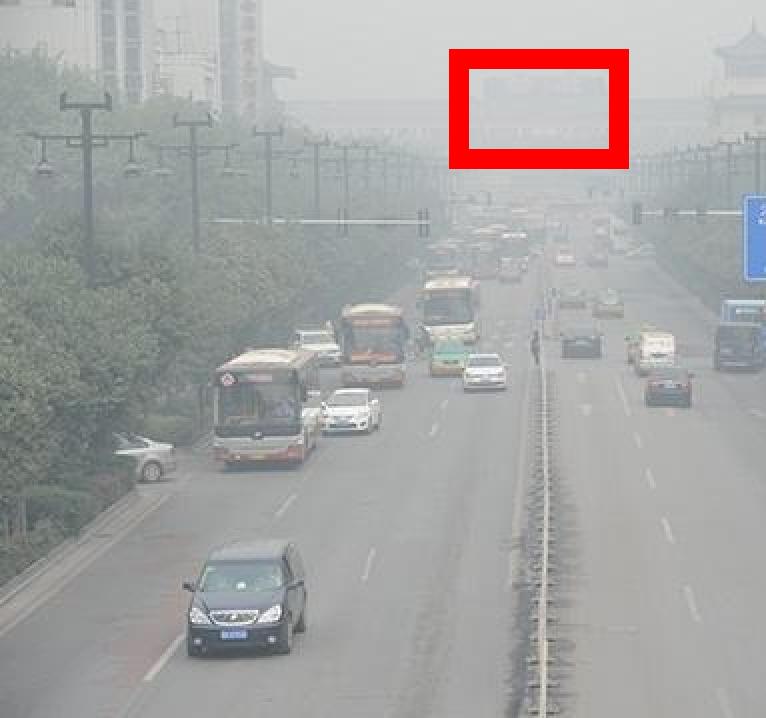}& 
        	\includegraphics[width=.3\linewidth]{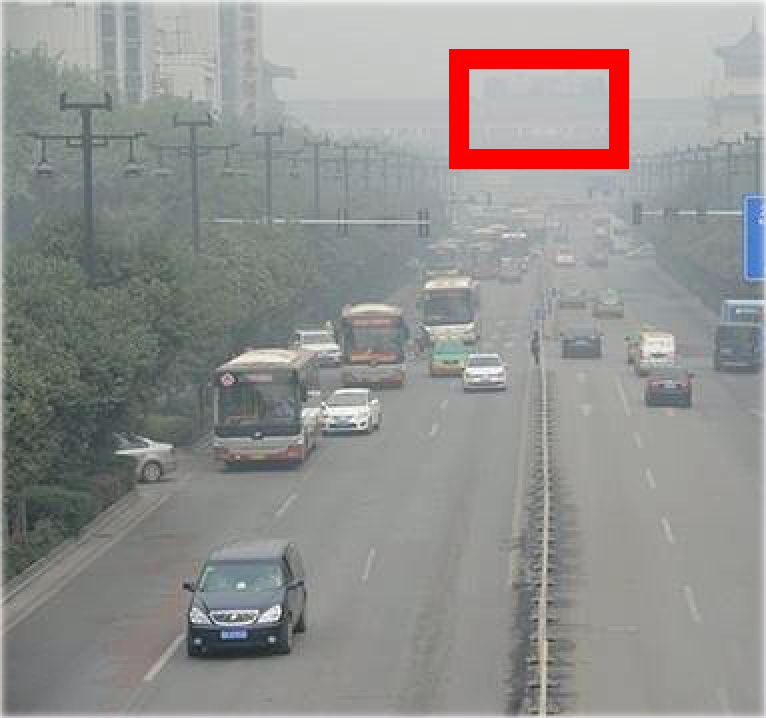}&
			\includegraphics[width=.3\linewidth]{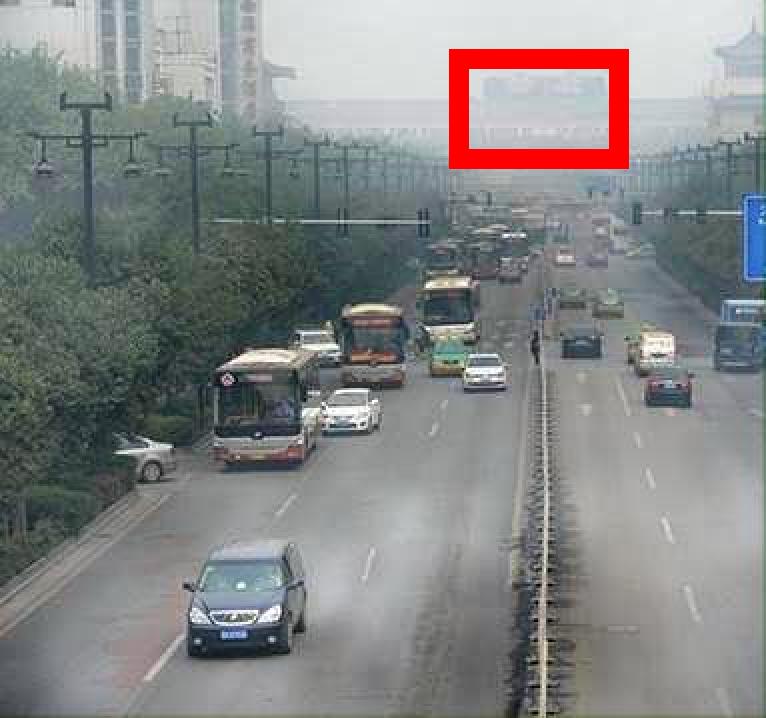}&
			\includegraphics[width=.3\linewidth]{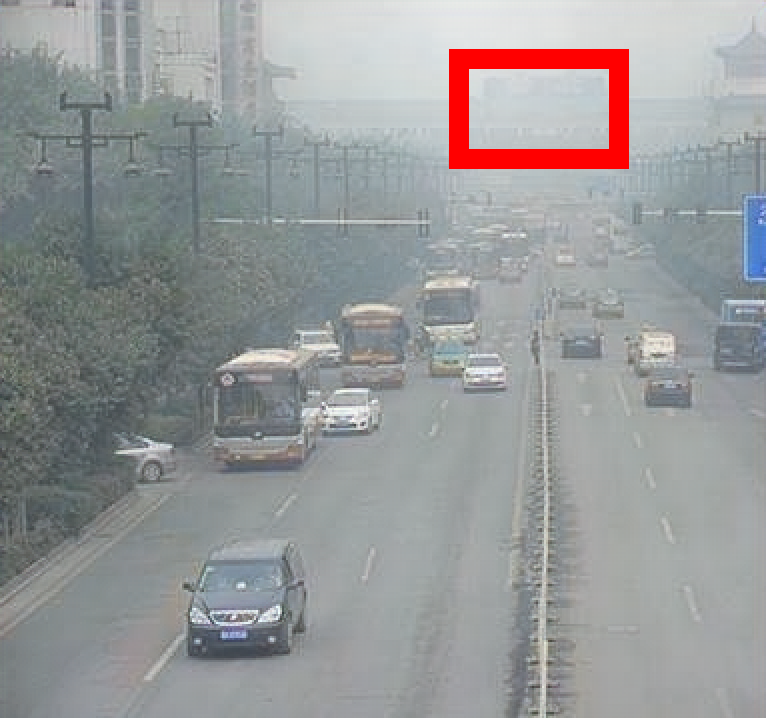}&
			\includegraphics[width=.3\linewidth]{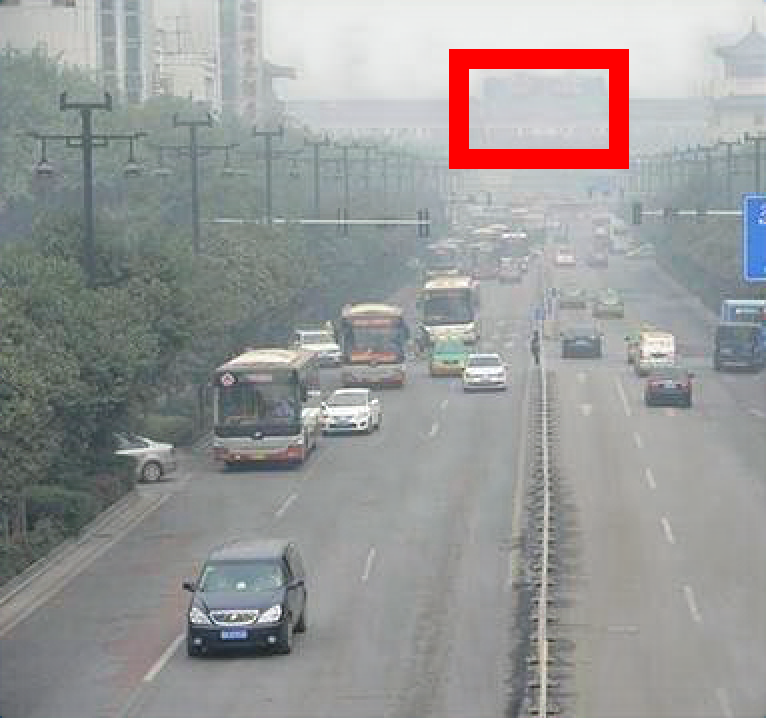} &
			\includegraphics[width=.3\linewidth]{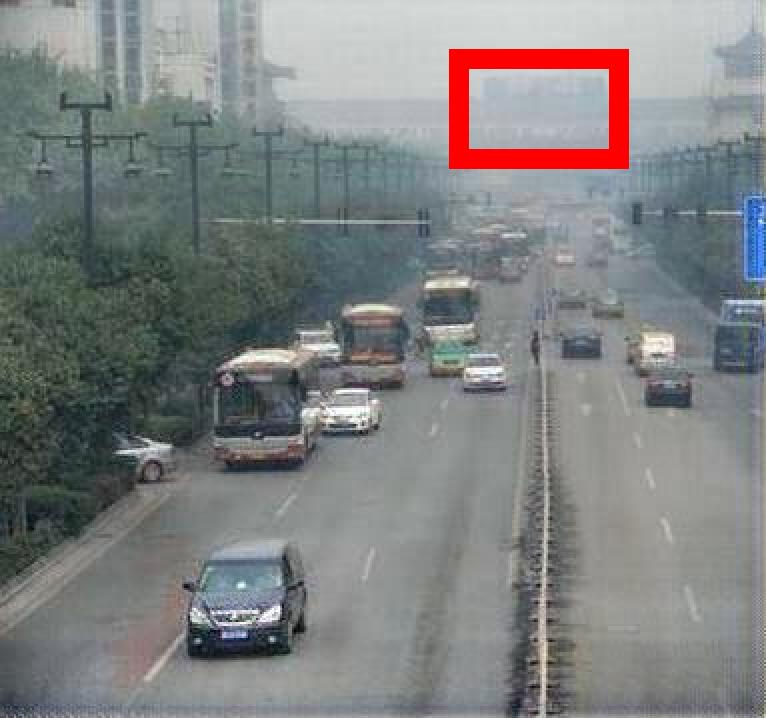}&
			\includegraphics[width=.3\linewidth]{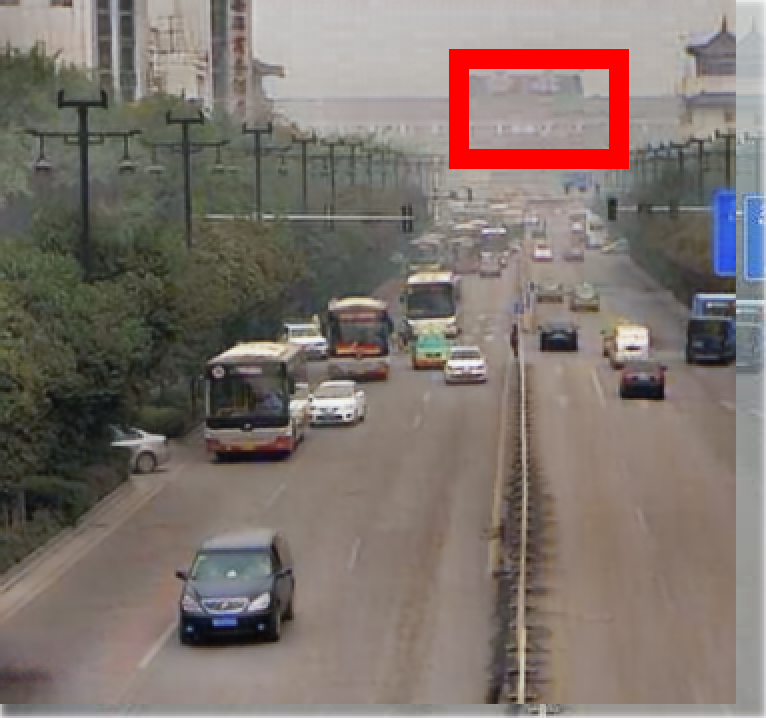}&
			\includegraphics[width=.3\linewidth]{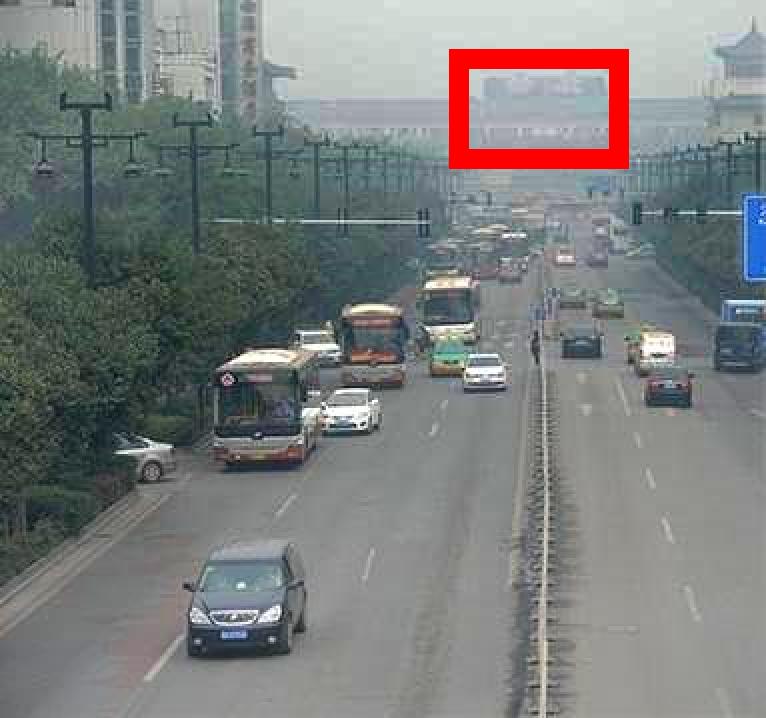}\\
			
			\includegraphics[width=.3\linewidth]{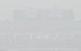}& 
        	\includegraphics[width=.3\linewidth]{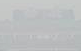}&
			\includegraphics[width=.3\linewidth]{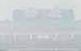}&
			\includegraphics[width=.3\linewidth]{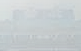}&
			\includegraphics[width=.3\linewidth]{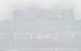} &
			\includegraphics[width=.3\linewidth]{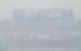}&
			\includegraphics[width=.3\linewidth]{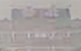}&
			\includegraphics[width=.3\linewidth]{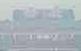}\\

			\includegraphics[width=.3\linewidth]{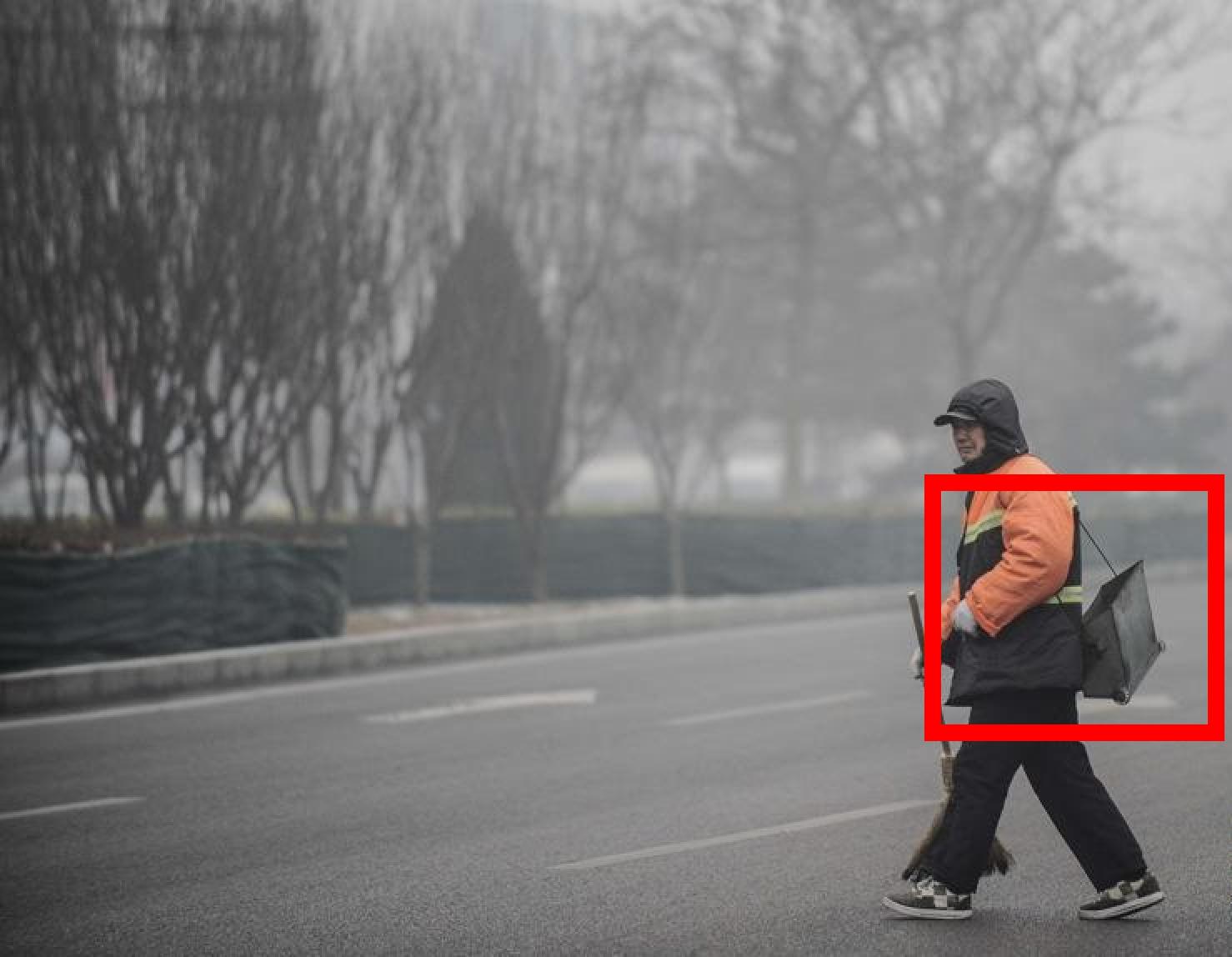}& 
        	\includegraphics[width=.3\linewidth]{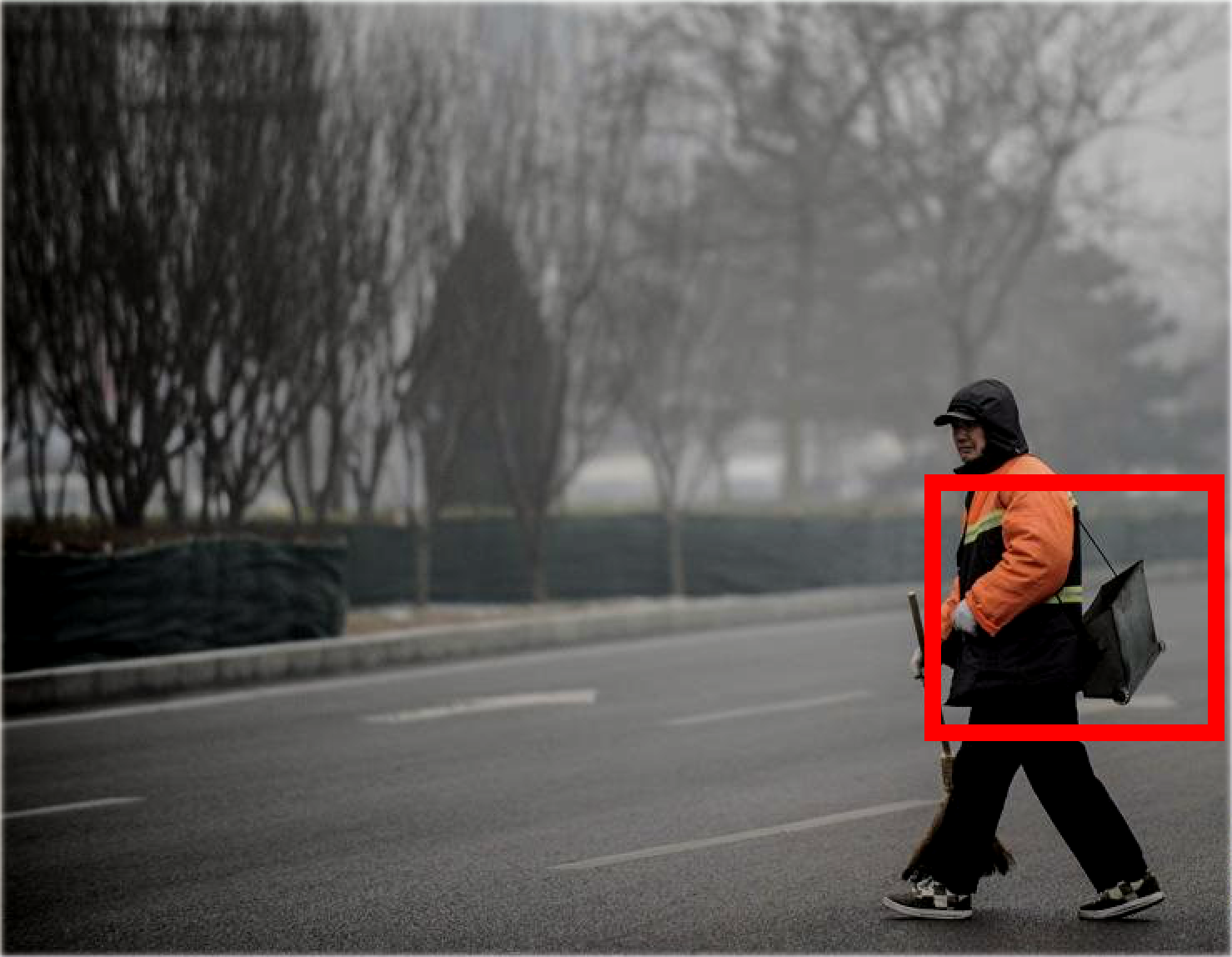}&
			\includegraphics[width=.3\linewidth]{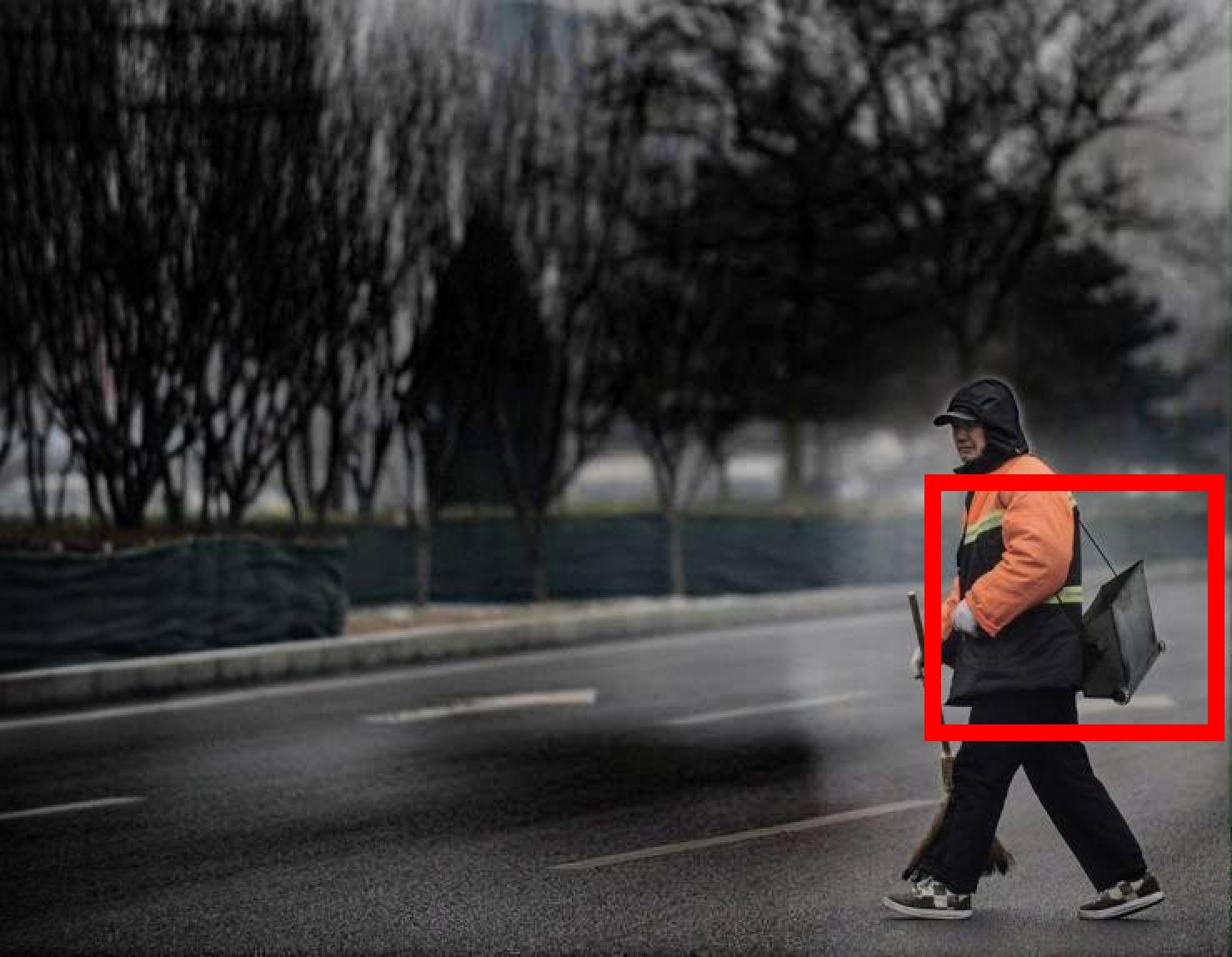}&
			\includegraphics[width=.3\linewidth]{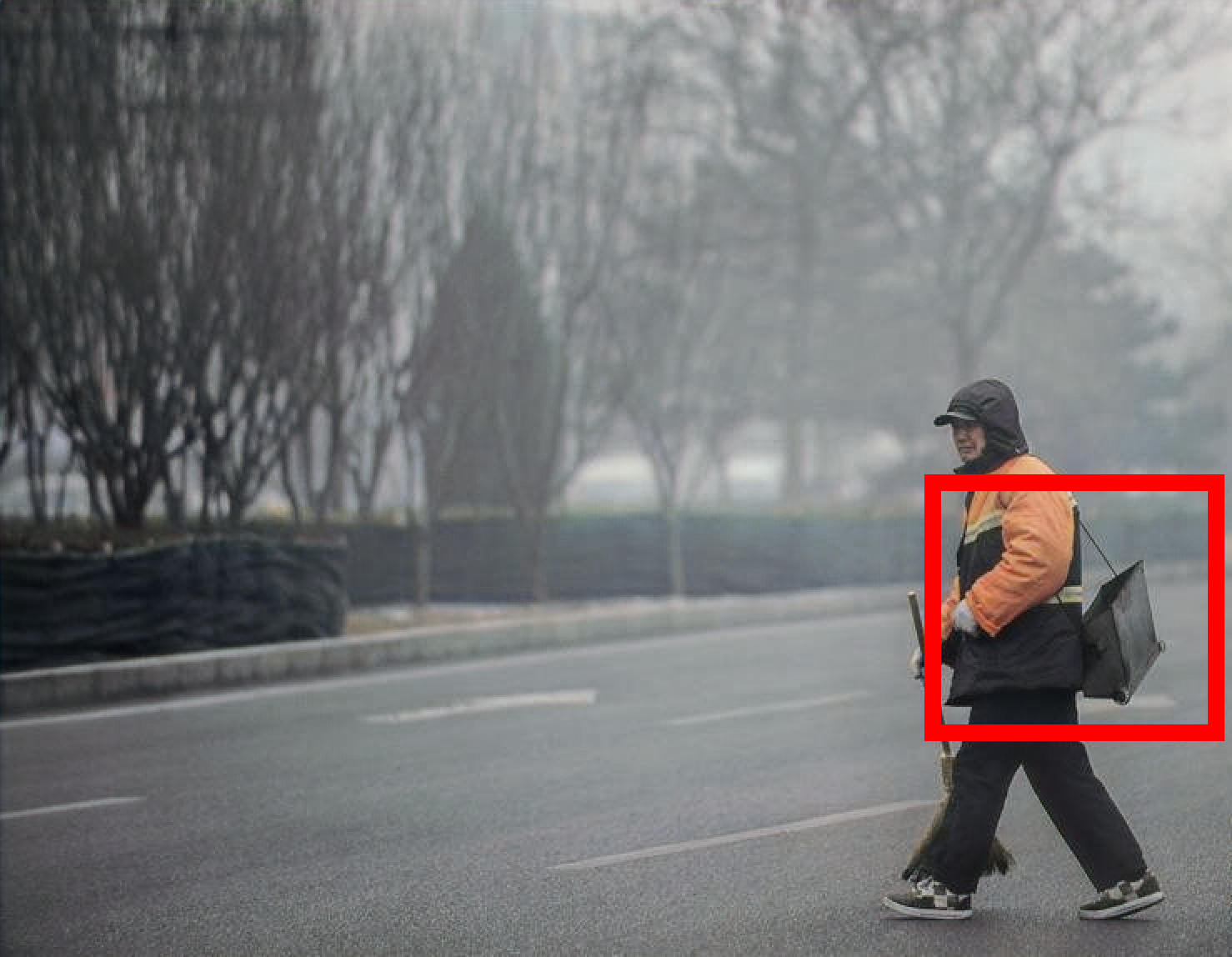}&
			\includegraphics[width=.3\linewidth]{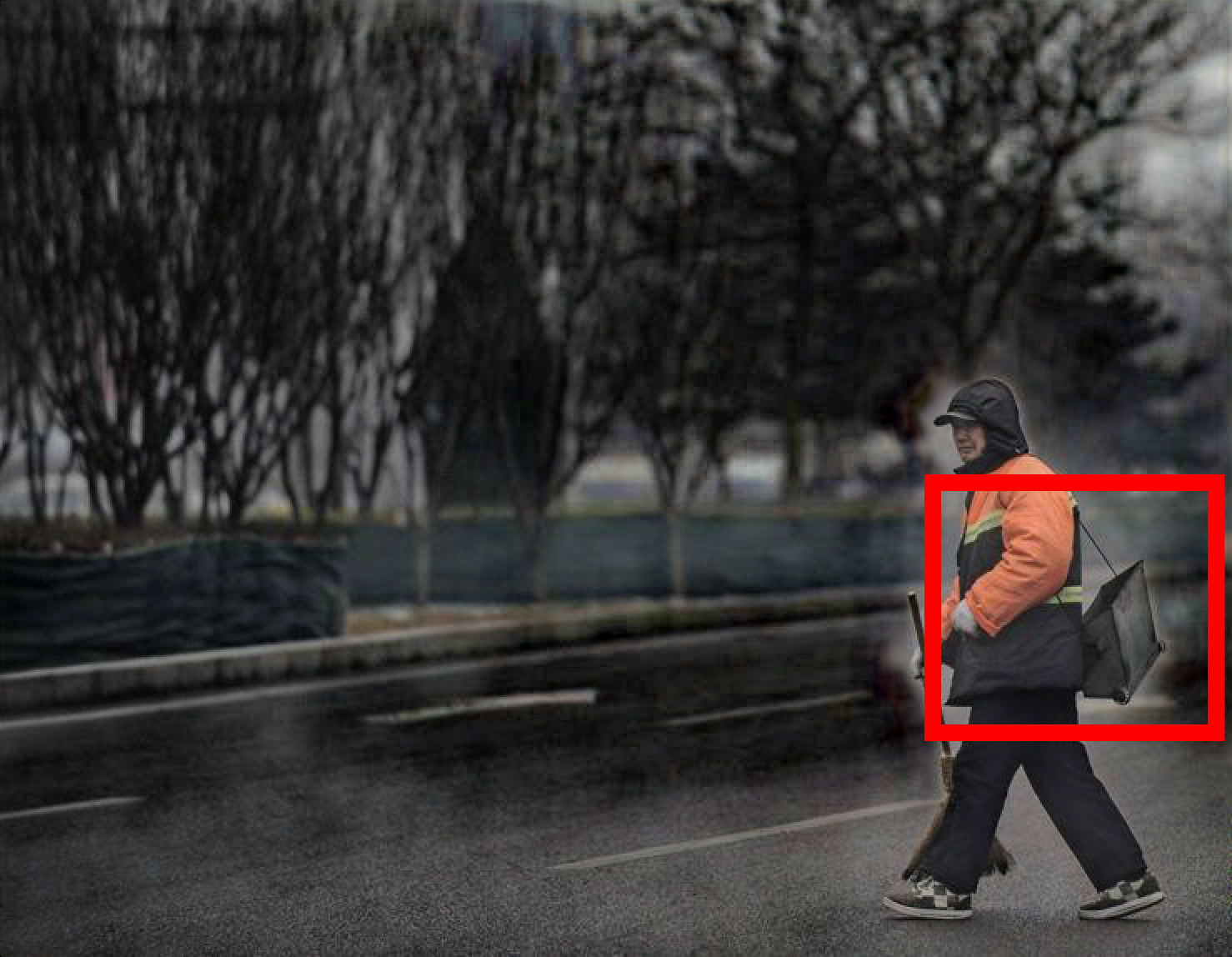} &
			\includegraphics[width=.3\linewidth]{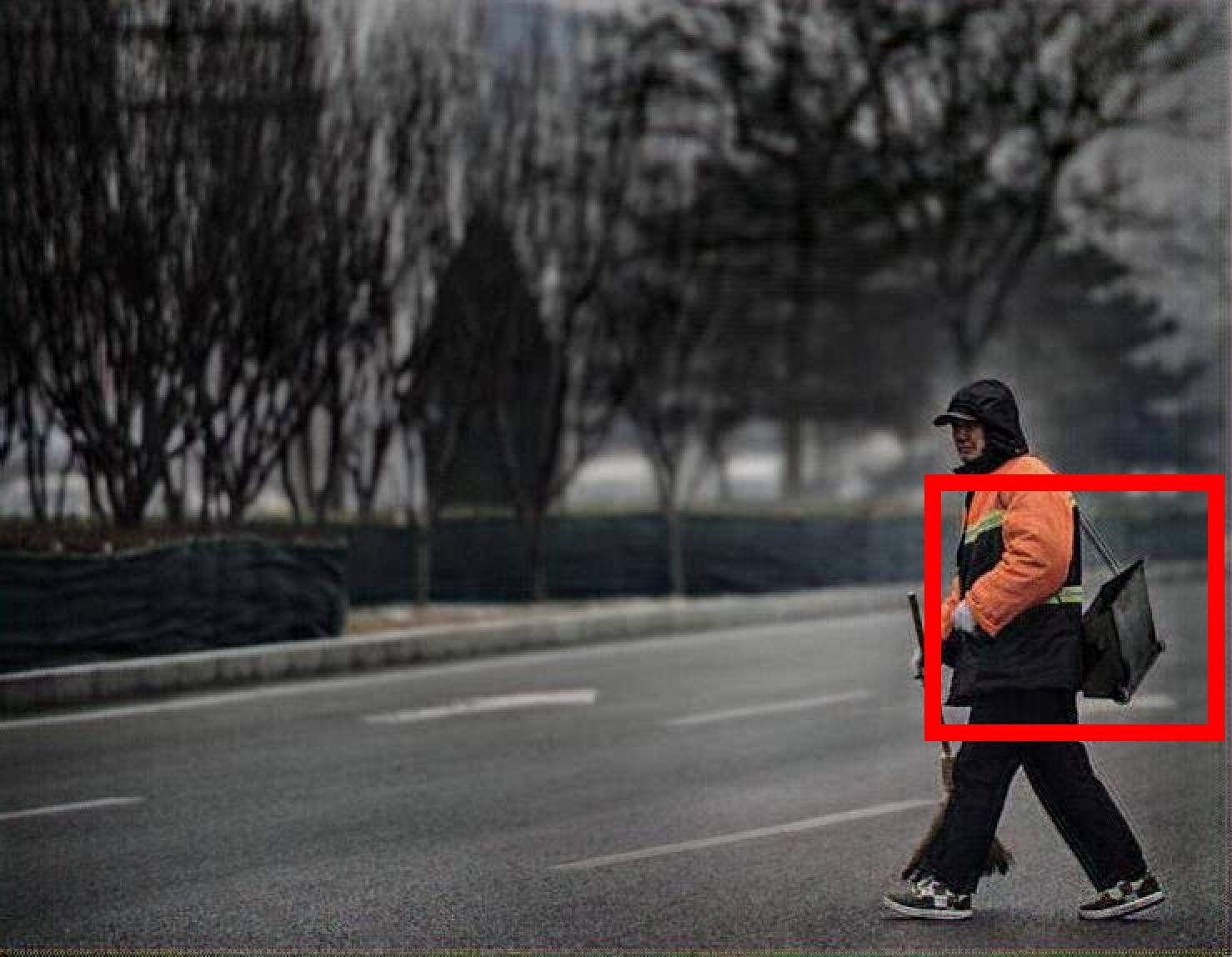}&
			\includegraphics[width=.3\linewidth]{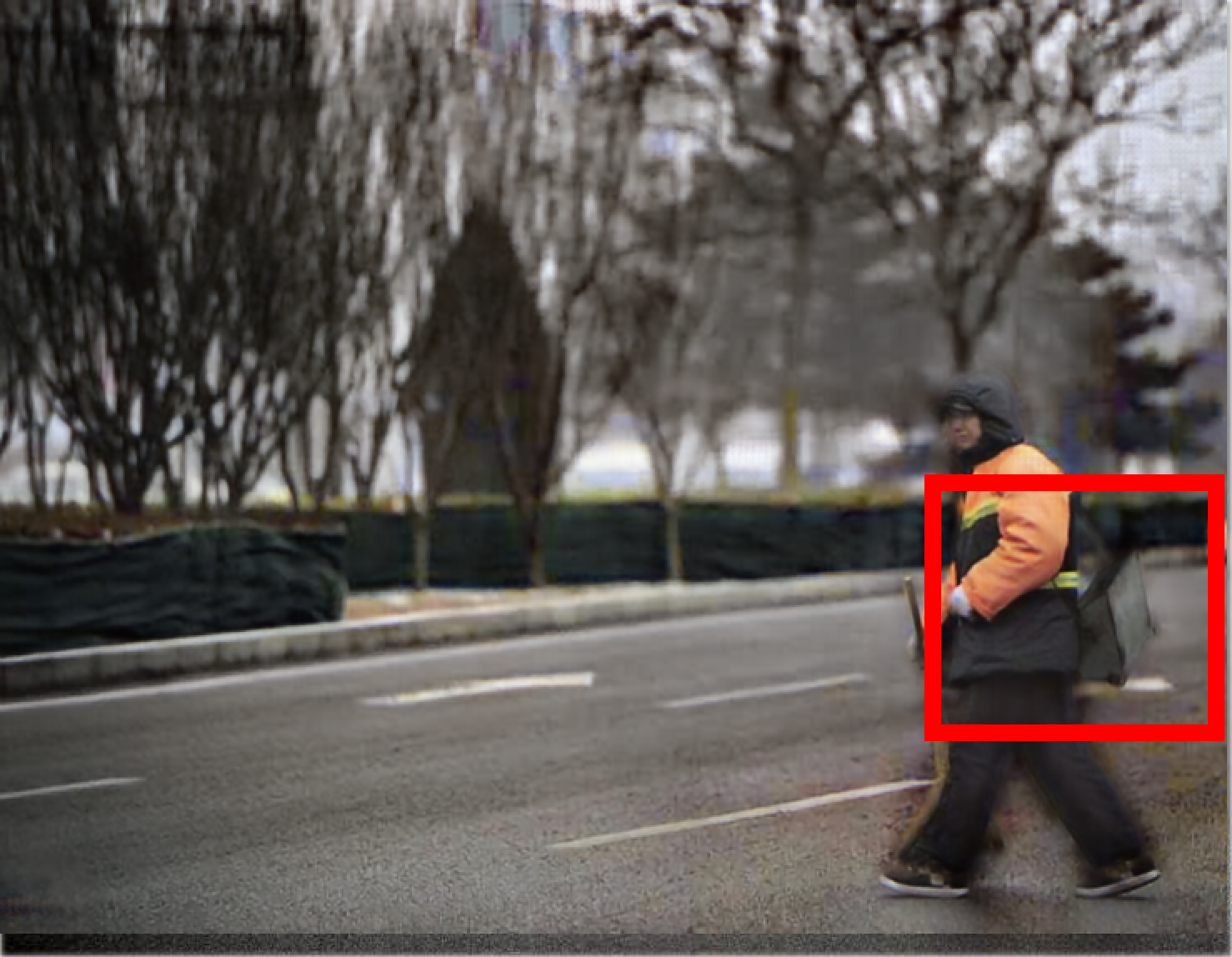}&
			\includegraphics[width=.3\linewidth]{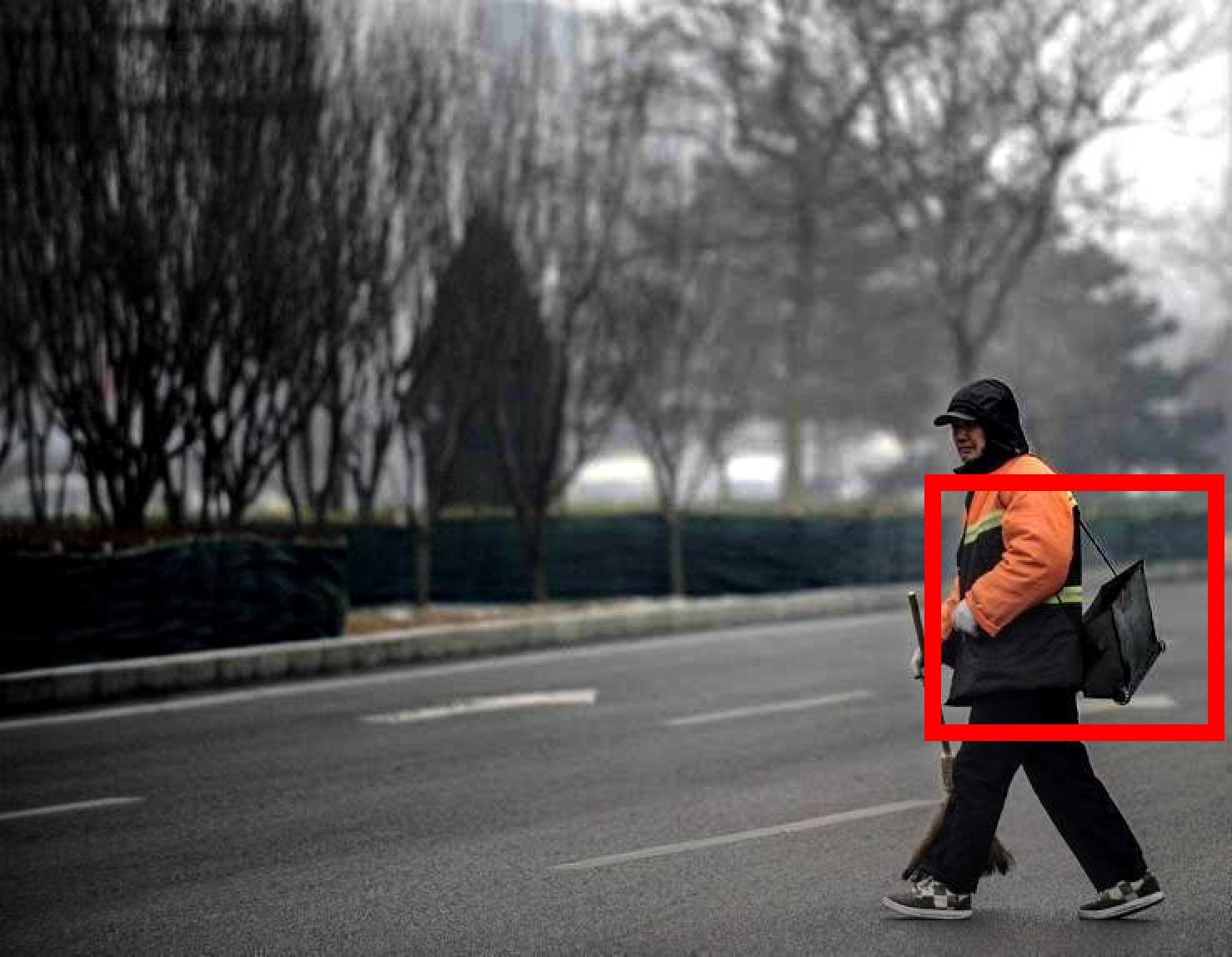}\\
			
			\includegraphics[width=.3\linewidth]{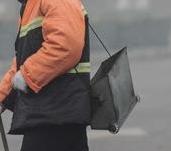}& 
        	\includegraphics[width=.3\linewidth]{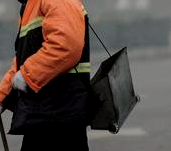}&
			\includegraphics[width=.3\linewidth]{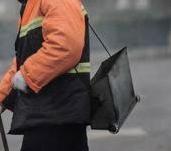}&
			\includegraphics[width=.3\linewidth]{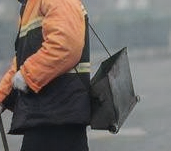}&
			\includegraphics[width=.3\linewidth]{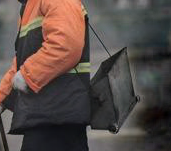} &
			\includegraphics[width=.3\linewidth]{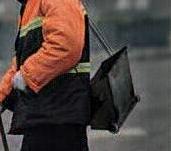}&
			\includegraphics[width=.3\linewidth]{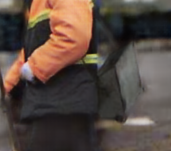}&
			\includegraphics[width=.3\linewidth]{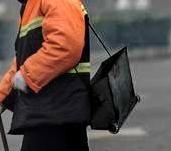}\\
			
			\Huge\raisebox{-0.2\height}{(a) Input Image} &
			\Huge\raisebox{-0.2\height}{(b) AOD-Net} &
			\Huge\raisebox{-0.2\height}{(c) GDN}&
			\Huge\raisebox{-0.2\height}{(d) DM$^2$F-Net }&
			\Huge\raisebox{-0.2\height}{(e) FFA-Net}&
			\Huge\raisebox{-0.2\height}{(f) MSBDN}&
			\Huge\raisebox{-0.2\height}{(g) DA}&
			\Huge\raisebox{-0.2\height}{(h) Our Method}\\
		\end{tabular}}
		\vspace{-3mm}
		\caption{Visual comparisons on dehazed results produced by our network (h) and SOTA methods (b)-(g) on more real-world hazy photos (a). Please see blown-up views for better visual comparisons. 
		}
		\label{fig:comparisons_real_part2}
		\vspace{-3mm}
	\end{figure*}

%% file: result-AB-figure.tex
\begin{figure*}	[t]
	\centering 
	\resizebox{0.9\linewidth}{!}{
		\setlength{\tabcolsep}{1.35mm}
		\begin{tabular}{cccccc}
			\Huge \raisebox{0.2\height}{PSNR / SSIM} &
			\Huge \raisebox{0.2\height}{21.65 / 0.81} &
			\Huge \raisebox{0.2\height}{26.72 / 0.83}&
			\Huge \raisebox{0.2\height}{27.43/0.92}&
			\Huge \raisebox{0.2\height}{29.64 / 0.94}&
			\Huge \raisebox{0.2\height}{$\infty$ / 1}\\
			
			\includegraphics[width=.3\linewidth]{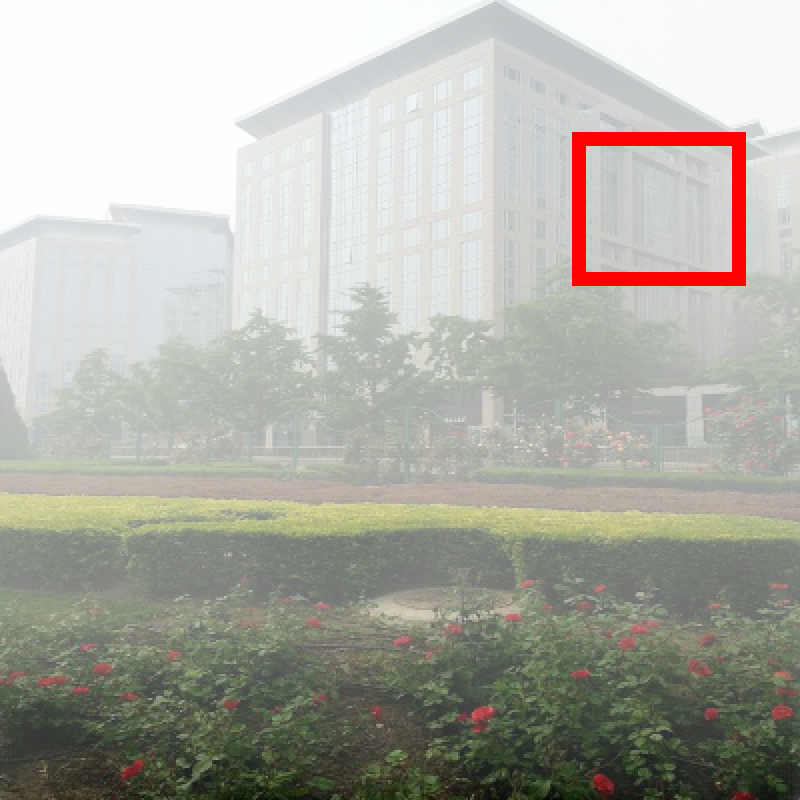}& 
        	\includegraphics[width=.3\linewidth]{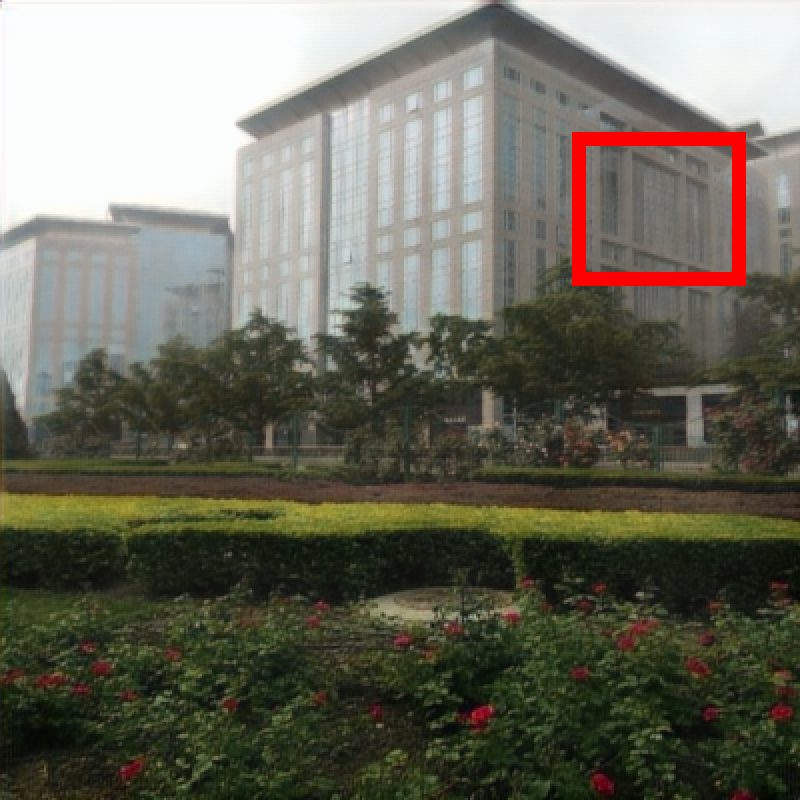}&
			\includegraphics[width=.3\linewidth]{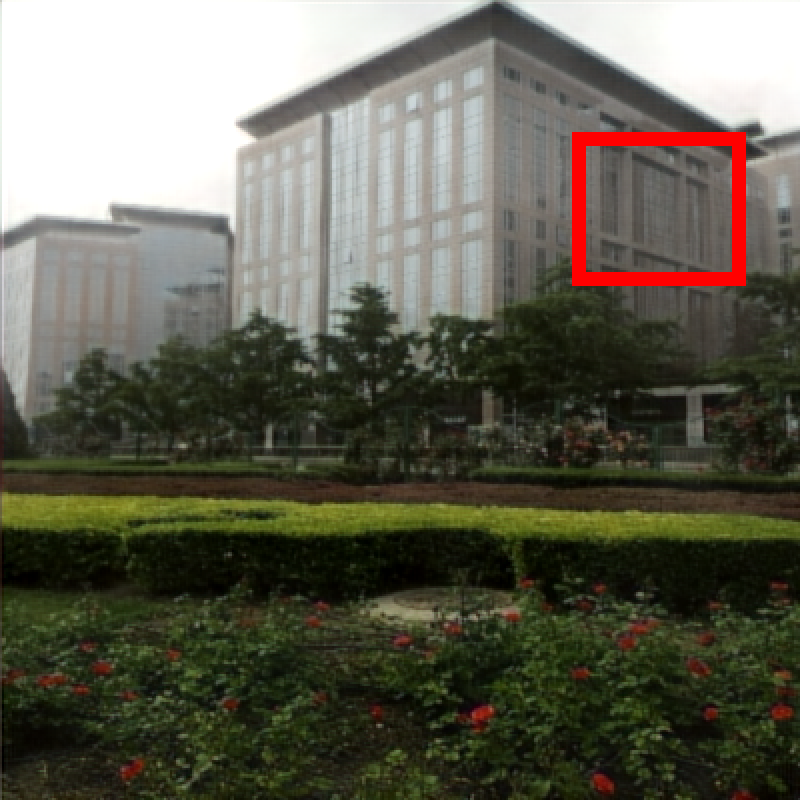}&
			\includegraphics[width=.3\linewidth]{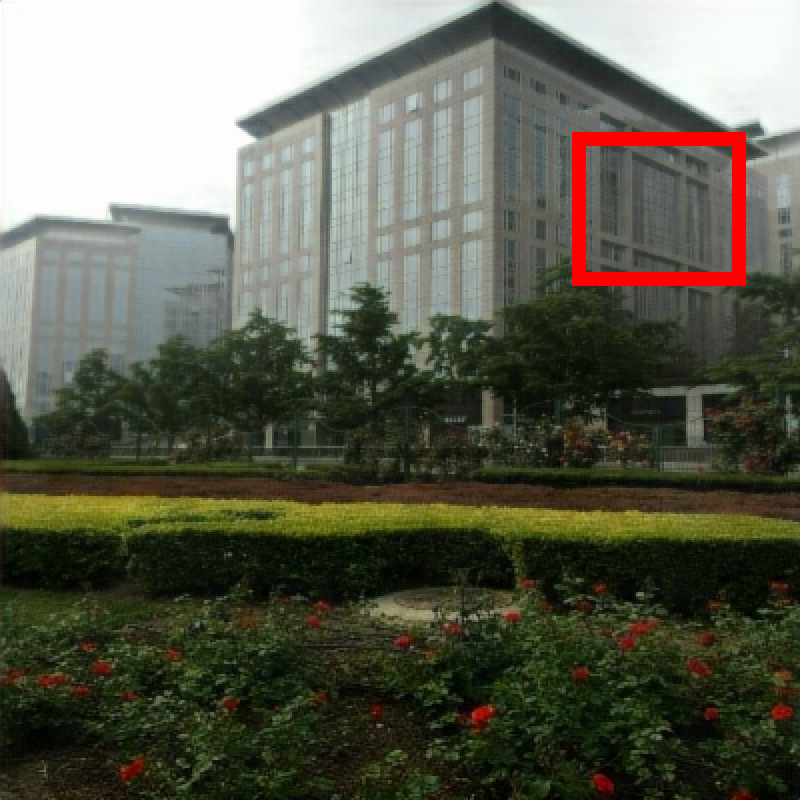}& 
			\includegraphics[width=.3\linewidth]{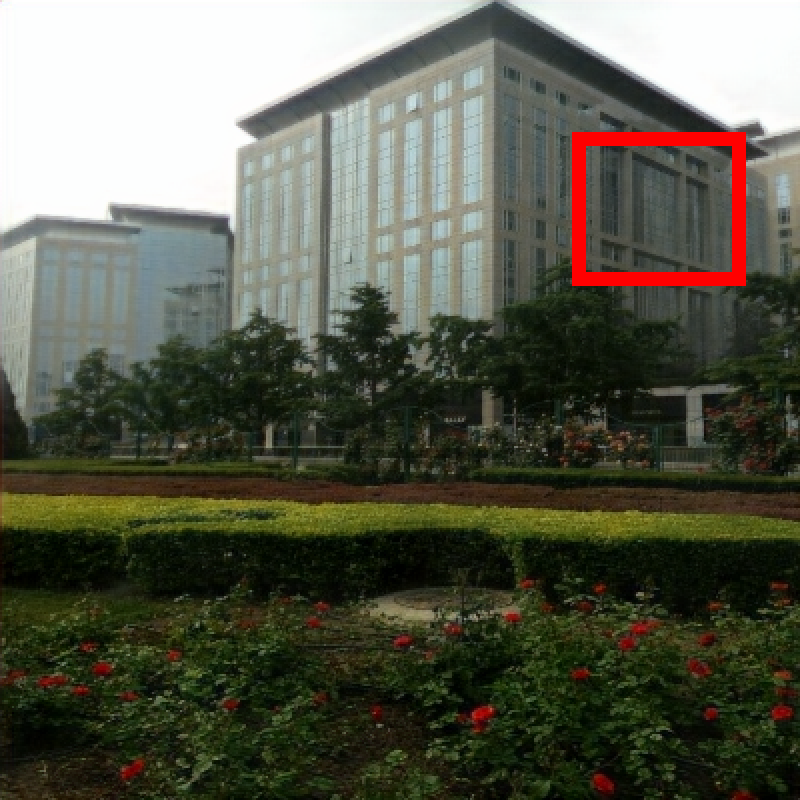} &
			\includegraphics[width=.3\linewidth]{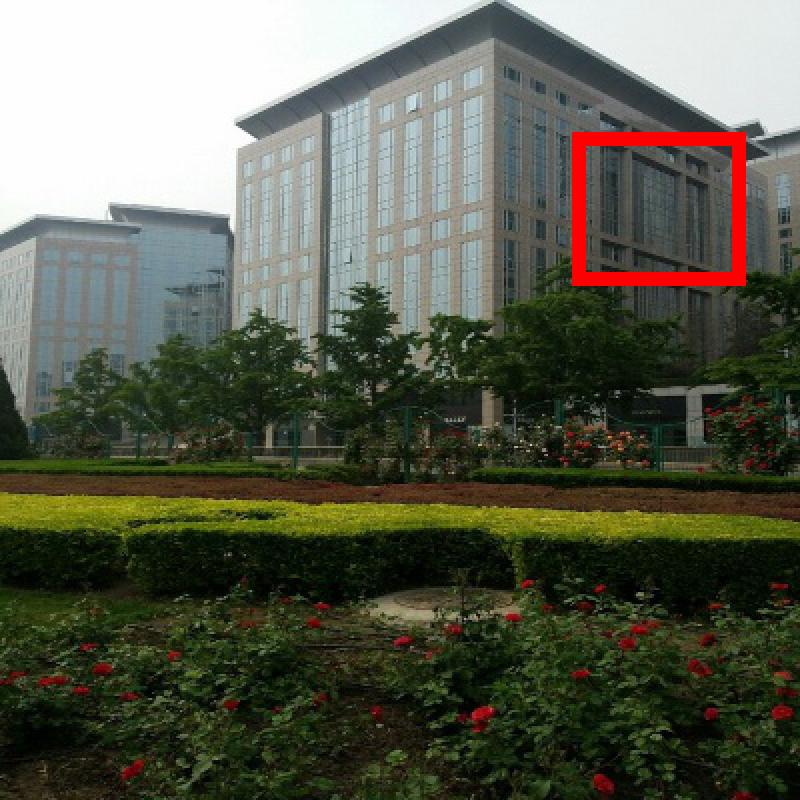}\\

			
			\includegraphics[width=.3\linewidth]{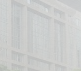}& 
			\includegraphics[width=.3\linewidth]{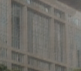}&
			\includegraphics[width=.3\linewidth]{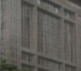}&
			\includegraphics[width=.3\linewidth]{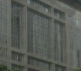}& 
			\includegraphics[width=.3\linewidth]{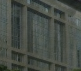} &
			\includegraphics[width=.3\linewidth]{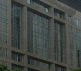}\\

			\Huge\raisebox{-0.2\height}{(a) Input Image} &
			\Huge\raisebox{-0.2\height}{(b) basic} &
			\Huge\raisebox{-0.2\height}{(c) basic+StageI}&
			\Huge\raisebox{-0.2\height}{(d) baisc+two-stages }&
			\Huge\raisebox{-0.2\height}{(e) Our method}&
			\Huge\raisebox{-0.2\height}{(f) Ground truth}\\
		\end{tabular}}
		\vspace{-3mm}
		\caption{Visual comparisons of dehazed results produced by our method and three baselines (ablation study). Please zoom in for a better illustration.
		}
		\label{fig:result-AB-figure}
		\vspace{-5mm}
	\end{figure*}

%% file: DMT-Net.bbl

\begin{thebibliography}{50}


\ifx \showCODEN    \undefined \def \showCODEN     #1{\unskip}     \fi
\ifx \showDOI      \undefined \def \showDOI       #1{#1}\fi
\ifx \showISBNx    \undefined \def \showISBNx     #1{\unskip}     \fi
\ifx \showISBNxiii \undefined \def \showISBNxiii  #1{\unskip}     \fi
\ifx \showISSN     \undefined \def \showISSN      #1{\unskip}     \fi
\ifx \showLCCN     \undefined \def \showLCCN      #1{\unskip}     \fi
\ifx \shownote     \undefined \def \shownote      #1{#1}          \fi
\ifx \showarticletitle \undefined \def \showarticletitle #1{#1}   \fi
\ifx \showURL      \undefined \def \showURL       {\relax}        \fi
\providecommand\bibfield[2]{#2}
\providecommand\bibinfo[2]{#2}
\providecommand\natexlab[1]{#1}
\providecommand\showeprint[2][]{arXiv:#2}

\bibitem[\protect\citeauthoryear{Ancuti and Ancuti}{Ancuti and Ancuti}{2013}]%
        {ancuti2013single}
\bibfield{author}{\bibinfo{person}{Codruta~Orniana Ancuti} {and}
  \bibinfo{person}{Cosmin Ancuti}.} \bibinfo{year}{2013}\natexlab{}.
\newblock \showarticletitle{Single image dehazing by multi-scale fusion}.
\newblock \bibinfo{journal}{\emph{TIP}} \bibinfo{volume}{22},
  \bibinfo{number}{8} (\bibinfo{year}{2013}), \bibinfo{pages}{3271--3282}.
\newblock


\bibitem[\protect\citeauthoryear{Berman and Avidan}{Berman and Avidan}{2016}]%
        {berman2016non}
\bibfield{author}{\bibinfo{person}{Dana Berman} {and} \bibinfo{person}{Shai
  Avidan}.} \bibinfo{year}{2016}\natexlab{}.
\newblock \showarticletitle{Non-local image dehazing}. In
  \bibinfo{booktitle}{\emph{CVPR}}. \bibinfo{pages}{1674--1682}.
\newblock


\bibitem[\protect\citeauthoryear{Berman, Treibitz, and Avidan}{Berman
  et~al\mbox{.}}{2017}]%
        {berman2017air}
\bibfield{author}{\bibinfo{person}{Dana Berman}, \bibinfo{person}{Tali
  Treibitz}, {and} \bibinfo{person}{Shai Avidan}.}
  \bibinfo{year}{2017}\natexlab{}.
\newblock \showarticletitle{Air-light estimation using haze-lines}. In
  \bibinfo{booktitle}{\emph{ICCP}}. \bibinfo{pages}{115--123}.
\newblock


\bibitem[\protect\citeauthoryear{Cai, Xu, Jia, Qing, and Tao}{Cai
  et~al\mbox{.}}{2016}]%
        {cai2016dehazenet}
\bibfield{author}{\bibinfo{person}{Bolun Cai}, \bibinfo{person}{Xiangmin Xu},
  \bibinfo{person}{Kui Jia}, \bibinfo{person}{Chunmei Qing}, {and}
  \bibinfo{person}{Dacheng Tao}.} \bibinfo{year}{2016}\natexlab{}.
\newblock \showarticletitle{Dehaze{N}et: An end-to-end system for single image
  haze removal}.
\newblock \bibinfo{journal}{\emph{TIP}} \bibinfo{volume}{25},
  \bibinfo{number}{11} (\bibinfo{year}{2016}), \bibinfo{pages}{5187--5198}.
\newblock


\bibitem[\protect\citeauthoryear{Chen, Zhu, Wan, Wang, Feng, and Heng}{Chen
  et~al\mbox{.}}{2020}]%
        {chen2020multi}
\bibfield{author}{\bibinfo{person}{Zhihao Chen}, \bibinfo{person}{Lei Zhu},
  \bibinfo{person}{Liang Wan}, \bibinfo{person}{Song Wang},
  \bibinfo{person}{Wei Feng}, {and} \bibinfo{person}{Pheng-Ann Heng}.}
  \bibinfo{year}{2020}\natexlab{}.
\newblock \showarticletitle{A Multi-task Mean Teacher for Semi-supervised
  Shadow Detection}. In \bibinfo{booktitle}{\emph{CVPR}}.
  \bibinfo{pages}{5610--5619}.
\newblock


\bibitem[\protect\citeauthoryear{Cheng, You, Ila, and Li}{Cheng
  et~al\mbox{.}}{2018}]%
        {cheng2018semantic}
\bibfield{author}{\bibinfo{person}{Ziang Cheng}, \bibinfo{person}{Shaodi You},
  \bibinfo{person}{Viorela Ila}, {and} \bibinfo{person}{Hongdong Li}.}
  \bibinfo{year}{2018}\natexlab{}.
\newblock \showarticletitle{Semantic single-image dehazing}.
\newblock \bibinfo{journal}{\emph{arXiv preprint arXiv:1804.05624}}
  (\bibinfo{year}{2018}).
\newblock


\bibitem[\protect\citeauthoryear{Deng, Huang, Tsai, and Lin}{Deng
  et~al\mbox{.}}{2020}]%
        {deng2020hardgan}
\bibfield{author}{\bibinfo{person}{Qili Deng}, \bibinfo{person}{Ziling Huang},
  \bibinfo{person}{Chung-Chi Tsai}, {and} \bibinfo{person}{Chia-Wen Lin}.}
  \bibinfo{year}{2020}\natexlab{}.
\newblock \showarticletitle{HardGAN: A Haze-Aware Representation Distillation
  GAN for Single Image Dehazing}. In \bibinfo{booktitle}{\emph{ECCV}}.
  \bibinfo{pages}{722--738}.
\newblock


\bibitem[\protect\citeauthoryear{Deng, Zhu, Hu, Fu, Xu, Zhang, Qin, and
  Heng}{Deng et~al\mbox{.}}{2019}]%
        {deng2019deep}
\bibfield{author}{\bibinfo{person}{Zijun Deng}, \bibinfo{person}{Lei Zhu},
  \bibinfo{person}{Xiaowei Hu}, \bibinfo{person}{Chi-Wing Fu},
  \bibinfo{person}{Xuemiao Xu}, \bibinfo{person}{Qing Zhang},
  \bibinfo{person}{Jing Qin}, {and} \bibinfo{person}{Pheng-Ann Heng}.}
  \bibinfo{year}{2019}\natexlab{}.
\newblock \showarticletitle{Deep multi-model fusion for single-image dehazing}.
  In \bibinfo{booktitle}{\emph{ICCV}}. \bibinfo{pages}{2453--2462}.
\newblock


\bibitem[\protect\citeauthoryear{Dong and Pan}{Dong and Pan}{2020}]%
        {dong2020physics}
\bibfield{author}{\bibinfo{person}{Jiangxin Dong} {and}
  \bibinfo{person}{Jinshan Pan}.} \bibinfo{year}{2020}\natexlab{}.
\newblock \showarticletitle{Physics-Based Feature Dehazing Networks}. In
  \bibinfo{booktitle}{\emph{ECCV}}. \bibinfo{pages}{188--204}.
\newblock


\bibitem[\protect\citeauthoryear{Fattal}{Fattal}{2008}]%
        {fattal2008single}
\bibfield{author}{\bibinfo{person}{Raanan Fattal}.}
  \bibinfo{year}{2008}\natexlab{}.
\newblock \showarticletitle{Single image dehazing}.
\newblock \bibinfo{journal}{\emph{TOGSIG}} \bibinfo{volume}{27},
  \bibinfo{number}{3} (\bibinfo{year}{2008}), \bibinfo{pages}{72:1--10}.
\newblock


\bibitem[\protect\citeauthoryear{Fattal}{Fattal}{2014}]%
        {fattal2014dehazing}
\bibfield{author}{\bibinfo{person}{Raanan Fattal}.}
  \bibinfo{year}{2014}\natexlab{}.
\newblock \showarticletitle{Dehazing using color-lines}.
\newblock \bibinfo{journal}{\emph{TOGSIG}} \bibinfo{volume}{34},
  \bibinfo{number}{1} (\bibinfo{year}{2014}), \bibinfo{pages}{13:1--14}.
\newblock


\bibitem[\protect\citeauthoryear{Galdran, Alvarez-Gila, Bria, Vazquez-Corral,
  and Bertalm{\i}o}{Galdran et~al\mbox{.}}{2018}]%
        {galdran2018duality}
\bibfield{author}{\bibinfo{person}{Adrian Galdran}, \bibinfo{person}{Aitor
  Alvarez-Gila}, \bibinfo{person}{Alessandro Bria}, \bibinfo{person}{Javier
  Vazquez-Corral}, {and} \bibinfo{person}{Marcelo Bertalm{\i}o}.}
  \bibinfo{year}{2018}\natexlab{}.
\newblock \showarticletitle{On the duality between {R}etinex and image
  dehazing}. In \bibinfo{booktitle}{\emph{CVPR}}. \bibinfo{pages}{8212--8221}.
\newblock


\bibitem[\protect\citeauthoryear{Hang, Jinshan, Zhe, Xiang, Xinyi, Fei, and
  Ming-Hsuan}{Hang et~al\mbox{.}}{2020}]%
        {MSBDN-DFF}
\bibfield{author}{\bibinfo{person}{Dong Hang}, \bibinfo{person}{Pan Jinshan},
  \bibinfo{person}{Hu Zhe}, \bibinfo{person}{Lei Xiang}, \bibinfo{person}{Zhang
  Xinyi}, \bibinfo{person}{Wang Fei}, {and} \bibinfo{person}{Yang Ming-Hsuan}.}
  \bibinfo{year}{2020}\natexlab{}.
\newblock \showarticletitle{Multi-Scale Boosted Dehazing Network with Dense
  Feature Fusion}. In \bibinfo{booktitle}{\emph{CVPR}}.
  \bibinfo{pages}{2154--2164}.
\newblock


\bibitem[\protect\citeauthoryear{He, Sun, and Tang}{He et~al\mbox{.}}{2011}]%
        {he2011single}
\bibfield{author}{\bibinfo{person}{Kaiming He}, \bibinfo{person}{Jian Sun},
  {and} \bibinfo{person}{Xiaoou Tang}.} \bibinfo{year}{2011}\natexlab{}.
\newblock \showarticletitle{Single image haze removal using dark channel
  prior}.
\newblock \bibinfo{journal}{\emph{TPAMI}} \bibinfo{volume}{33},
  \bibinfo{number}{12} (\bibinfo{year}{2011}), \bibinfo{pages}{2341--2353}.
\newblock


\bibitem[\protect\citeauthoryear{Isola, Zhu, Zhou, and Efros}{Isola
  et~al\mbox{.}}{2017}]%
        {isola2017image}
\bibfield{author}{\bibinfo{person}{Phillip Isola}, \bibinfo{person}{Jun-Yan
  Zhu}, \bibinfo{person}{Tinghui Zhou}, {and} \bibinfo{person}{Alexei~A
  Efros}.} \bibinfo{year}{2017}\natexlab{}.
\newblock \showarticletitle{Image-to-image translation with conditional
  adversarial networks}. In \bibinfo{booktitle}{\emph{CVPR}}.
  \bibinfo{pages}{5967--5976}.
\newblock


\bibitem[\protect\citeauthoryear{Laine and Aila}{Laine and Aila}{2016}]%
        {laine2016temporal}
\bibfield{author}{\bibinfo{person}{Samuli Laine} {and} \bibinfo{person}{Timo
  Aila}.} \bibinfo{year}{2016}\natexlab{}.
\newblock \showarticletitle{Temporal ensembling for semi-supervised learning}.
\newblock \bibinfo{journal}{\emph{arXiv}} (\bibinfo{year}{2016}).
\newblock


\bibitem[\protect\citeauthoryear{Li, Gou, Liu, Zhu, Zhou, and Peng}{Li
  et~al\mbox{.}}{2020a}]%
        {li2020zero}
\bibfield{author}{\bibinfo{person}{Boyun Li}, \bibinfo{person}{Yuanbiao Gou},
  \bibinfo{person}{Jerry~Zitao Liu}, \bibinfo{person}{Hongyuan Zhu},
  \bibinfo{person}{Joey~Tianyi Zhou}, {and} \bibinfo{person}{Xi Peng}.}
  \bibinfo{year}{2020}\natexlab{a}.
\newblock \showarticletitle{Zero-shot image dehazing}.
\newblock \bibinfo{journal}{\emph{TIP}}  \bibinfo{volume}{29}
  (\bibinfo{year}{2020}), \bibinfo{pages}{8457--8466}.
\newblock


\bibitem[\protect\citeauthoryear{Li, Peng, Wang, Xu, and Feng}{Li
  et~al\mbox{.}}{2017}]%
        {li2017all}
\bibfield{author}{\bibinfo{person}{Boyi Li}, \bibinfo{person}{Xiulian Peng},
  \bibinfo{person}{Zhangyang Wang}, \bibinfo{person}{Jizheng Xu}, {and}
  \bibinfo{person}{Dan Feng}.} \bibinfo{year}{2017}\natexlab{}.
\newblock \showarticletitle{{AOD}-{N}et: An all-in-one network for dehazing and
  beyond}. In \bibinfo{booktitle}{\emph{ICCV}}.
\newblock


\bibitem[\protect\citeauthoryear{Li, Ren, Fu, Tao, Feng, Zeng, and Wang}{Li
  et~al\mbox{.}}{2018c}]%
        {li2018benchmarking}
\bibfield{author}{\bibinfo{person}{Boyi Li}, \bibinfo{person}{Wenqi Ren},
  \bibinfo{person}{Dengpan Fu}, \bibinfo{person}{Dacheng Tao},
  \bibinfo{person}{Dan Feng}, \bibinfo{person}{Wenjun Zeng}, {and}
  \bibinfo{person}{Zhangyang Wang}.} \bibinfo{year}{2018}\natexlab{c}.
\newblock \showarticletitle{Benchmarking single-image dehazing and beyond}.
\newblock \bibinfo{journal}{\emph{TIP}} \bibinfo{volume}{28},
  \bibinfo{number}{1} (\bibinfo{year}{2018}), \bibinfo{pages}{492--505}.
\newblock


\bibitem[\protect\citeauthoryear{Li, Ren, Fu, Tao, Feng, Zeng, and Wang}{Li
  et~al\mbox{.}}{2019}]%
        {li2019benchmarking}
\bibfield{author}{\bibinfo{person}{Boyi Li}, \bibinfo{person}{Wenqi Ren},
  \bibinfo{person}{Dengpan Fu}, \bibinfo{person}{Dacheng Tao},
  \bibinfo{person}{Dan Feng}, \bibinfo{person}{Wenjun Zeng}, {and}
  \bibinfo{person}{Zhangyang Wang}.} \bibinfo{year}{2019}\natexlab{}.
\newblock \showarticletitle{Benchmarking Single-Image Dehazing and Beyond}.
\newblock \bibinfo{journal}{\emph{TIP}} \bibinfo{volume}{28},
  \bibinfo{number}{1} (\bibinfo{year}{2019}), \bibinfo{pages}{492--505}.
\newblock


\bibitem[\protect\citeauthoryear{Li, Guo, Guo, Han, Fu, and Cong}{Li
  et~al\mbox{.}}{2020b}]%
        {Li2020_tmm}
\bibfield{author}{\bibinfo{person}{Chongyi Li}, \bibinfo{person}{Chunle Guo},
  \bibinfo{person}{Jichang Guo}, \bibinfo{person}{Ping Han},
  \bibinfo{person}{Huazhu Fu}, {and} \bibinfo{person}{Runmin Cong}.}
  \bibinfo{year}{2020}\natexlab{b}.
\newblock \showarticletitle{{PDR-Net: Perception-Inspired Single Image Dehazing
  Network With Refinement}}.
\newblock \bibinfo{journal}{\emph{TMM}} \bibinfo{volume}{22},
  \bibinfo{number}{3} (\bibinfo{year}{2020}), \bibinfo{pages}{704--716}.
\newblock


\bibitem[\protect\citeauthoryear{Li, Guo, Porikli, Fu, and Pang}{Li
  et~al\mbox{.}}{2018a}]%
        {Li2018_Access}
\bibfield{author}{\bibinfo{person}{Chongyi Li}, \bibinfo{person}{Jichang Guo},
  \bibinfo{person}{Fatih Porikli}, \bibinfo{person}{Huazhu Fu}, {and}
  \bibinfo{person}{Yanwei Pang}.} \bibinfo{year}{2018}\natexlab{a}.
\newblock \showarticletitle{{A Cascaded Convolutional Neural Network for Single
  Image Dehazing}}.
\newblock \bibinfo{journal}{\emph{IEEE Access}}  \bibinfo{volume}{6}
  (\bibinfo{year}{2018}), \bibinfo{pages}{24877--24887}.
\newblock
\showISSN{2169-3536}


\bibitem[\protect\citeauthoryear{Li, Pan, Li, and Tang}{Li
  et~al\mbox{.}}{2018b}]%
        {Li_2018_CVPR}
\bibfield{author}{\bibinfo{person}{Runde Li}, \bibinfo{person}{Jinshan Pan},
  \bibinfo{person}{Zechao Li}, {and} \bibinfo{person}{Jinhui Tang}.}
  \bibinfo{year}{2018}\natexlab{b}.
\newblock \showarticletitle{Single Image Dehazing via Conditional Generative
  Adversarial Network}. In \bibinfo{booktitle}{\emph{CVPR}}.
  \bibinfo{pages}{8202--8211}.
\newblock


\bibitem[\protect\citeauthoryear{Liu, Rabinovich, and Berg}{Liu
  et~al\mbox{.}}{2016}]%
        {liu2015parsenet}
\bibfield{author}{\bibinfo{person}{Wei Liu}, \bibinfo{person}{Andrew
  Rabinovich}, {and} \bibinfo{person}{Alexander~C Berg}.}
  \bibinfo{year}{2016}\natexlab{}.
\newblock \showarticletitle{Parse{N}et: Looking wider to see better}. In
  \bibinfo{booktitle}{\emph{ICLR}}.
\newblock


\bibitem[\protect\citeauthoryear{Liu, Ma, Shi, and Chen}{Liu
  et~al\mbox{.}}{2019}]%
        {liu2019griddehazenet}
\bibfield{author}{\bibinfo{person}{Xiaohong Liu}, \bibinfo{person}{Yongrui Ma},
  \bibinfo{person}{Zhihao Shi}, {and} \bibinfo{person}{Jun Chen}.}
  \bibinfo{year}{2019}\natexlab{}.
\newblock \showarticletitle{Griddehazenet: Attention-based multi-scale network
  for image dehazing}. In \bibinfo{booktitle}{\emph{ICCV}}.
  \bibinfo{pages}{7313--7322}.
\newblock


\bibitem[\protect\citeauthoryear{Liu, Wei, Shao, Sheng, Yan, and Wang}{Liu
  et~al\mbox{.}}{2018}]%
        {liu2018exploring}
\bibfield{author}{\bibinfo{person}{Yu Liu}, \bibinfo{person}{Fangyin Wei},
  \bibinfo{person}{Jing Shao}, \bibinfo{person}{Lu Sheng},
  \bibinfo{person}{Junjie Yan}, {and} \bibinfo{person}{Xiaogang Wang}.}
  \bibinfo{year}{2018}\natexlab{}.
\newblock \showarticletitle{Exploring disentangled feature representation
  beyond face identification}. In \bibinfo{booktitle}{\emph{CVPR}}.
  \bibinfo{pages}{2080--2089}.
\newblock


\bibitem[\protect\citeauthoryear{Nayar and Narasimhan}{Nayar and
  Narasimhan}{1999}]%
        {nayar1999vision}
\bibfield{author}{\bibinfo{person}{Shree~K Nayar} {and}
  \bibinfo{person}{Srinivasa~G Narasimhan}.} \bibinfo{year}{1999}\natexlab{}.
\newblock \showarticletitle{Vision in bad weather}. In
  \bibinfo{booktitle}{\emph{ICCV}}. \bibinfo{pages}{820--827}.
\newblock


\bibitem[\protect\citeauthoryear{Qin, Wang, Bai, Xie, and Jia}{Qin
  et~al\mbox{.}}{2020}]%
        {2020FFA}
\bibfield{author}{\bibinfo{person}{Xu Qin}, \bibinfo{person}{Zhilin Wang},
  \bibinfo{person}{Yuanchao Bai}, \bibinfo{person}{Xiaodong Xie}, {and}
  \bibinfo{person}{Huizhu Jia}.} \bibinfo{year}{2020}\natexlab{}.
\newblock \showarticletitle{FFA-Net: Feature Fusion Attention Network for
  Single Image Dehazing}. In \bibinfo{booktitle}{\emph{AAAI}}.
  \bibinfo{pages}{11908--11915}.
\newblock


\bibitem[\protect\citeauthoryear{Qu, Chen, Huang, and Xie}{Qu
  et~al\mbox{.}}{2019}]%
        {qu2019enhanced}
\bibfield{author}{\bibinfo{person}{Yanyun Qu}, \bibinfo{person}{Yizi Chen},
  \bibinfo{person}{Jingying Huang}, {and} \bibinfo{person}{Yuan Xie}.}
  \bibinfo{year}{2019}\natexlab{}.
\newblock \showarticletitle{Enhanced pix2pix dehazing network}. In
  \bibinfo{booktitle}{\emph{CVPR}}. \bibinfo{pages}{8160--8168}.
\newblock


\bibitem[\protect\citeauthoryear{Ren, Liu, Zhang, Pan, Cao, and Yang}{Ren
  et~al\mbox{.}}{2016}]%
        {ren2016single}
\bibfield{author}{\bibinfo{person}{Wenqi Ren}, \bibinfo{person}{Si Liu},
  \bibinfo{person}{Hua Zhang}, \bibinfo{person}{Jinshan Pan},
  \bibinfo{person}{Xiaochun Cao}, {and} \bibinfo{person}{Ming-Hsuan Yang}.}
  \bibinfo{year}{2016}\natexlab{}.
\newblock \showarticletitle{Single image dehazing via multi-scale convolutional
  neural networks}. In \bibinfo{booktitle}{\emph{ECCV}}.
  \bibinfo{pages}{154--169}.
\newblock


\bibitem[\protect\citeauthoryear{Ren, Ma, Zhang, Pan, Cao, Liu, and Yang}{Ren
  et~al\mbox{.}}{2018}]%
        {ren2018gated}
\bibfield{author}{\bibinfo{person}{Wenqi Ren}, \bibinfo{person}{Lin Ma},
  \bibinfo{person}{Jiawei Zhang}, \bibinfo{person}{Jinshan Pan},
  \bibinfo{person}{Xiaochun Cao}, \bibinfo{person}{Wei Liu}, {and}
  \bibinfo{person}{Ming-Hsuan Yang}.} \bibinfo{year}{2018}\natexlab{}.
\newblock \showarticletitle{Gated fusion network for single image dehazing}. In
  \bibinfo{booktitle}{\emph{CVPR}}. \bibinfo{pages}{3253--3261}.
\newblock


\bibitem[\protect\citeauthoryear{Ronneberger, Fischer, and Brox}{Ronneberger
  et~al\mbox{.}}{2015}]%
        {ronneberger2015u}
\bibfield{author}{\bibinfo{person}{Olaf Ronneberger}, \bibinfo{person}{Philipp
  Fischer}, {and} \bibinfo{person}{Thomas Brox}.}
  \bibinfo{year}{2015}\natexlab{}.
\newblock \showarticletitle{U-net: Convolutional networks for biomedical image
  segmentation}. In \bibinfo{booktitle}{\emph{MICCAI}}.
  \bibinfo{pages}{234--241}.
\newblock


\bibitem[\protect\citeauthoryear{Shao, Li, Ren, Gao, and Sang}{Shao
  et~al\mbox{.}}{2020}]%
        {shao2020domain}
\bibfield{author}{\bibinfo{person}{Yuanjie Shao}, \bibinfo{person}{Lerenhan
  Li}, \bibinfo{person}{Wenqi Ren}, \bibinfo{person}{Changxin Gao}, {and}
  \bibinfo{person}{Nong Sang}.} \bibinfo{year}{2020}\natexlab{}.
\newblock \showarticletitle{Domain Adaptation for Image Dehazing}. In
  \bibinfo{booktitle}{\emph{CVPR}}. \bibinfo{pages}{2805--2814}.
\newblock


\bibitem[\protect\citeauthoryear{Silberman, Hoiem, Kohli, and Fergus}{Silberman
  et~al\mbox{.}}{2012}]%
        {silberman2012indoor}
\bibfield{author}{\bibinfo{person}{Nathan Silberman}, \bibinfo{person}{Derek
  Hoiem}, \bibinfo{person}{Pushmeet Kohli}, {and} \bibinfo{person}{Rob
  Fergus}.} \bibinfo{year}{2012}\natexlab{}.
\newblock \showarticletitle{Indoor segmentation and support inference from rgbd
  images}. In \bibinfo{booktitle}{\emph{ECCV}}. \bibinfo{pages}{746--760}.
\newblock


\bibitem[\protect\citeauthoryear{Simonyan and Zisserman}{Simonyan and
  Zisserman}{2015}]%
        {simonyan2015very}
\bibfield{author}{\bibinfo{person}{Karen Simonyan} {and}
  \bibinfo{person}{Andrew Zisserman}.} \bibinfo{year}{2015}\natexlab{}.
\newblock \showarticletitle{Very deep convolutional networks for large-scale
  image recognition}. In \bibinfo{booktitle}{\emph{ICLR}}.
\newblock


\bibitem[\protect\citeauthoryear{Song, Li, Wang, and Chen}{Song
  et~al\mbox{.}}{2018}]%
        {2018Single}
\bibfield{author}{\bibinfo{person}{Yafei Song}, \bibinfo{person}{Jia Li},
  \bibinfo{person}{Xiaogang Wang}, {and} \bibinfo{person}{Xiaowu Chen}.}
  \bibinfo{year}{2018}\natexlab{}.
\newblock \showarticletitle{Single Image Dehazing Using Ranking Convolutional
  Neural Network}.
\newblock \bibinfo{journal}{\emph{TMM}} \bibinfo{volume}{20},
  \bibinfo{number}{6} (\bibinfo{year}{2018}), \bibinfo{pages}{1548--1560}.
\newblock


\bibitem[\protect\citeauthoryear{Sulami, Glatzer, Fattal, and Werman}{Sulami
  et~al\mbox{.}}{2014}]%
        {sulami2014automatic}
\bibfield{author}{\bibinfo{person}{Matan Sulami}, \bibinfo{person}{Itamar
  Glatzer}, \bibinfo{person}{Raanan Fattal}, {and} \bibinfo{person}{Mike
  Werman}.} \bibinfo{year}{2014}\natexlab{}.
\newblock \showarticletitle{Automatic recovery of the atmospheric light in hazy
  images}. In \bibinfo{booktitle}{\emph{ICCP}}. \bibinfo{pages}{1--11}.
\newblock


\bibitem[\protect\citeauthoryear{Tarvainen and Valpola}{Tarvainen and
  Valpola}{2017}]%
        {tarvainen2017mean}
\bibfield{author}{\bibinfo{person}{Antti Tarvainen} {and}
  \bibinfo{person}{Harri Valpola}.} \bibinfo{year}{2017}\natexlab{}.
\newblock \showarticletitle{Mean teachers are better role models:
  Weight-averaged consistency targets improve semi-supervised deep learning
  results}. In \bibinfo{booktitle}{\emph{NIPS}}. \bibinfo{pages}{1195--1204}.
\newblock


\bibitem[\protect\citeauthoryear{Wang, Yuan, Wu, and Liu}{Wang
  et~al\mbox{.}}{2017}]%
        {2017Fast}
\bibfield{author}{\bibinfo{person}{Wencheng Wang}, \bibinfo{person}{Xiaohui
  Yuan}, \bibinfo{person}{Xiaojin Wu}, {and} \bibinfo{person}{Yunlong Liu}.}
  \bibinfo{year}{2017}\natexlab{}.
\newblock \showarticletitle{Fast Image Dehazing Method Based on Linear
  Transformation}.
\newblock \bibinfo{journal}{\emph{TMM}} \bibinfo{volume}{19},
  \bibinfo{number}{6} (\bibinfo{year}{2017}), \bibinfo{pages}{1142--1155}.
\newblock


\bibitem[\protect\citeauthoryear{Wang, Bovik, Sheikh, and Simoncelli}{Wang
  et~al\mbox{.}}{2004}]%
        {wang2004image}
\bibfield{author}{\bibinfo{person}{Zhou Wang}, \bibinfo{person}{Alan~C Bovik},
  \bibinfo{person}{Hamid~R Sheikh}, {and} \bibinfo{person}{Eero~P Simoncelli}.}
  \bibinfo{year}{2004}\natexlab{}.
\newblock \showarticletitle{Image quality assessment: from error visibility to
  structural similarity}.
\newblock \bibinfo{journal}{\emph{TIP}} \bibinfo{volume}{13},
  \bibinfo{number}{4} (\bibinfo{year}{2004}), \bibinfo{pages}{600--612}.
\newblock


\bibitem[\protect\citeauthoryear{Xie, Girshick, Doll{\'a}r, Tu, and He}{Xie
  et~al\mbox{.}}{2017}]%
        {xie2017aggregated}
\bibfield{author}{\bibinfo{person}{Saining Xie}, \bibinfo{person}{Ross
  Girshick}, \bibinfo{person}{Piotr Doll{\'a}r}, \bibinfo{person}{Zhuowen Tu},
  {and} \bibinfo{person}{Kaiming He}.} \bibinfo{year}{2017}\natexlab{}.
\newblock \showarticletitle{Aggregated residual transformations for deep neural
  networks}. In \bibinfo{booktitle}{\emph{CVPR}}. \bibinfo{pages}{5987--5995}.
\newblock


\bibitem[\protect\citeauthoryear{Yang and Sun}{Yang and Sun}{2018}]%
        {yang2018proximal}
\bibfield{author}{\bibinfo{person}{Dong Yang} {and} \bibinfo{person}{Jian
  Sun}.} \bibinfo{year}{2018}\natexlab{}.
\newblock \showarticletitle{Proximal Dehaze-Net: A Prior Learning-Based Deep
  Network for Single Image Dehazing}. In \bibinfo{booktitle}{\emph{ECCV}}.
  \bibinfo{pages}{729--746}.
\newblock


\bibitem[\protect\citeauthoryear{Zhang and Patel}{Zhang and Patel}{2018}]%
        {zhang2018densely}
\bibfield{author}{\bibinfo{person}{He Zhang} {and} \bibinfo{person}{Vishal~M
  Patel}.} \bibinfo{year}{2018}\natexlab{}.
\newblock \showarticletitle{Densely connected pyramid dehazing network}. In
  \bibinfo{booktitle}{\emph{CVPR}}. \bibinfo{pages}{3194--3203}.
\newblock


\bibitem[\protect\citeauthoryear{Zhang, Ding, and Sharma}{Zhang
  et~al\mbox{.}}{2017}]%
        {zhang2017hazerd}
\bibfield{author}{\bibinfo{person}{Yanfu Zhang}, \bibinfo{person}{Li Ding},
  {and} \bibinfo{person}{Gaurav Sharma}.} \bibinfo{year}{2017}\natexlab{}.
\newblock \showarticletitle{{HAZERD}: an outdoor scene dataset and benchmark
  for single image dehazing}. In \bibinfo{booktitle}{\emph{ICIP}}.
  \bibinfo{pages}{3205--3209}.
\newblock


\bibitem[\protect\citeauthoryear{Zhang, Li, Li, Wang, Zhong, and Fu}{Zhang
  et~al\mbox{.}}{2018}]%
        {zhang2018image}
\bibfield{author}{\bibinfo{person}{Yulun Zhang}, \bibinfo{person}{Kunpeng Li},
  \bibinfo{person}{Kai Li}, \bibinfo{person}{Lichen Wang},
  \bibinfo{person}{Bineng Zhong}, {and} \bibinfo{person}{Yun Fu}.}
  \bibinfo{year}{2018}\natexlab{}.
\newblock \showarticletitle{Image super-resolution using very deep residual
  channel attention networks}. In \bibinfo{booktitle}{\emph{ECCV}}.
  \bibinfo{pages}{294--310}.
\newblock


\bibitem[\protect\citeauthoryear{Zheng, Ren, Cao, Hu, Wang, Song, and
  Jia}{Zheng et~al\mbox{.}}{2021}]%
        {Zheng2021}
\bibfield{author}{\bibinfo{person}{Zhuoran Zheng}, \bibinfo{person}{Wenqi Ren},
  \bibinfo{person}{Xiaochun Cao}, \bibinfo{person}{Xiaobin Hu},
  \bibinfo{person}{Tao Wang}, \bibinfo{person}{Fenglong Song}, {and}
  \bibinfo{person}{Xiuyi Jia}.} \bibinfo{year}{2021}\natexlab{}.
\newblock \showarticletitle{Ultra-High-Definition Image Dehazing via
  Multi-Guided Bilateral Learning}. In \bibinfo{booktitle}{\emph{CVPR}}.
\newblock


\bibitem[\protect\citeauthoryear{Zhu, Deng, Hu, Xie, Xu, Qin, and Heng}{Zhu
  et~al\mbox{.}}{2021}]%
        {zhu2020learning}
\bibfield{author}{\bibinfo{person}{Lei Zhu}, \bibinfo{person}{Zijun Deng},
  \bibinfo{person}{Xiaowei Hu}, \bibinfo{person}{Haoran Xie},
  \bibinfo{person}{Xuemiao Xu}, \bibinfo{person}{Jing Qin}, {and}
  \bibinfo{person}{Pheng-Ann Heng}.} \bibinfo{year}{2021}\natexlab{}.
\newblock \showarticletitle{Learning gated non-local residual for single-image
  rain streak removal}.
\newblock \bibinfo{journal}{\emph{IEEE Transactions on Circuits and Systems for
  Video Technology}} \bibinfo{volume}{31}, \bibinfo{number}{6}
  (\bibinfo{year}{2021}), \bibinfo{pages}{2147--2159}.
\newblock


\bibitem[\protect\citeauthoryear{Zhu, Fu, Brown, and Heng}{Zhu
  et~al\mbox{.}}{2017a}]%
        {zhu2017non}
\bibfield{author}{\bibinfo{person}{Lei Zhu}, \bibinfo{person}{Chi-Wing Fu},
  \bibinfo{person}{Michael~S Brown}, {and} \bibinfo{person}{Pheng-Ann Heng}.}
  \bibinfo{year}{2017}\natexlab{a}.
\newblock \showarticletitle{A non-local low-rank framework for ultrasound
  speckle reduction}. In \bibinfo{booktitle}{\emph{CVPR}}.
  \bibinfo{pages}{493--501}.
\newblock


\bibitem[\protect\citeauthoryear{Zhu, Fu, Lischinski, and Heng}{Zhu
  et~al\mbox{.}}{2017b}]%
        {zhu2017joint}
\bibfield{author}{\bibinfo{person}{Lei Zhu}, \bibinfo{person}{Chi-Wing Fu},
  \bibinfo{person}{Dani Lischinski}, {and} \bibinfo{person}{Pheng-Ann Heng}.}
  \bibinfo{year}{2017}\natexlab{b}.
\newblock \showarticletitle{Joint bi-layer optimization for single-image rain
  streak removal}. In \bibinfo{booktitle}{\emph{Proceedings of the IEEE
  international conference on computer vision}}. \bibinfo{pages}{2526--2534}.
\newblock


\bibitem[\protect\citeauthoryear{Zhu, Mai, Shao, et~al\mbox{.}}{Zhu
  et~al\mbox{.}}{2015}]%
        {zhu2015fast}
\bibfield{author}{\bibinfo{person}{Qingsong Zhu}, \bibinfo{person}{Jiaming
  Mai}, \bibinfo{person}{Ling Shao}, {et~al\mbox{.}}}
  \bibinfo{year}{2015}\natexlab{}.
\newblock \showarticletitle{A Fast Single Image Haze Removal Algorithm Using
  Color Attenuation Prior.}
\newblock \bibinfo{journal}{\emph{TIP}} \bibinfo{volume}{24},
  \bibinfo{number}{11} (\bibinfo{year}{2015}), \bibinfo{pages}{3522--3533}.
\newblock


\end{thebibliography}
